\pdfoutput=1

\documentclass[11pt]{article}

\usepackage{acl}

\usepackage{times}
\usepackage{latexsym}

\usepackage[T1]{fontenc}

\usepackage[utf8]{inputenc}

\usepackage{microtype}

\usepackage[noend, ruled]{algorithm2e}
\usepackage{amssymb}
\usepackage{amsmath}
\usepackage{graphicx}
\usepackage{caption}
\usepackage{booktabs}
\usepackage{multirow}
\DeclareCaptionLabelFormat{andtable}{#1~#2  \&  \tablename~\thetable}
\newcommand{\taskname}{CFKGR}

\title{Counterfactual Reasoning with Knowledge Graph Embeddings}

\author{Lena Zellinger$^{1}$, Andreas Stephan$^{1,2, 4}$, \and Benjamin Roth$^{1,3}$ \\
$^1$ Faculty of Computer Science, University of Vienna, Austria \\ 
$^2$ UniVie Doctoral School Computer Science, University of Vienna, Austria \\
$^3$ Faculty of Philological and Cultural Studies, University of Vienna, Austria\\
\texttt{$^4${andreas.stephan}@univie.ac.at}}

\begin{document}
\maketitle
\begin{abstract}
Knowledge graph embeddings (KGEs) were originally developed to infer true but missing facts in incomplete knowledge repositories.
In this paper, we link knowledge graph completion and counterfactual reasoning via our new task \taskname. We model the original world state as a knowledge graph, hypothetical scenarios as edges added to the graph, and plausible changes to the graph as inferences from logical rules.
We create corresponding benchmark datasets, which contain diverse hypothetical scenarios with plausible changes to the original knowledge graph and facts that should be retained. 
We develop \emph{COULDD},
a general method for adapting existing knowledge graph embeddings given a hypothetical premise, and evaluate it on our benchmark.
Our results indicate that KGEs learn patterns in the graph without explicit training. We further observe that KGEs adapted with COULDD 
solidly 
detect plausible counterfactual changes to the graph that follow these patterns. An evaluation on human-annotated data reveals that KGEs adapted with COULDD are mostly unable to recognize changes to the graph that do not follow learned inference rules. In contrast, ChatGPT mostly
outperforms KGEs in detecting plausible changes to the graph but has poor knowledge retention. 
In summary, \taskname{} connects two previously distinct areas, namely KG completion and counterfactual reasoning.
\end{abstract}

\section{Introduction}\label{sec:intro}
Reasoning about hypothetical situations (\emph{counterfactual reasoning}) and anticipating the effects of a change in the current state of the world is central to human cognition \cite{rafetseder2014counterfactual, van2015cognitive}, and has been identified as a key concept in game theory \cite{aumann1995backward, halpern1999hypothetical} and agent-based systems \cite{icard2018instrumental,parvaneh2020counterfactual}.
It has even been argued that the capacity to reason about alternative configurations of the world could be a pre-requisite to the existence of free will and a sense of agency \cite{mccarthy2000free,kulakova2017could}. 
Recently, there has been an increased interest in evaluating and improving counterfactual reasoning of AI systems, in particular, large language models (LLMs) \cite{qin-etal-2019-counterfactual, frohberg2022crass, li-etal-2023-counterfactual}. 

Knowledge graphs (KGs) express rich information about the world as an explicit collection of triples, such as (Paris, capital, France), and knowledge graph embeddings (KGEs) effectively infer true but missing facts from incomplete knowledge repositories \cite{hogan2021knowledge,ji2021survey}.
Yet, to the best of our knowledge, KGEs have not been explored for counterfactual reasoning.
\begin{figure}[t!]
\begin{center}
\includegraphics[width=0.4\textwidth]{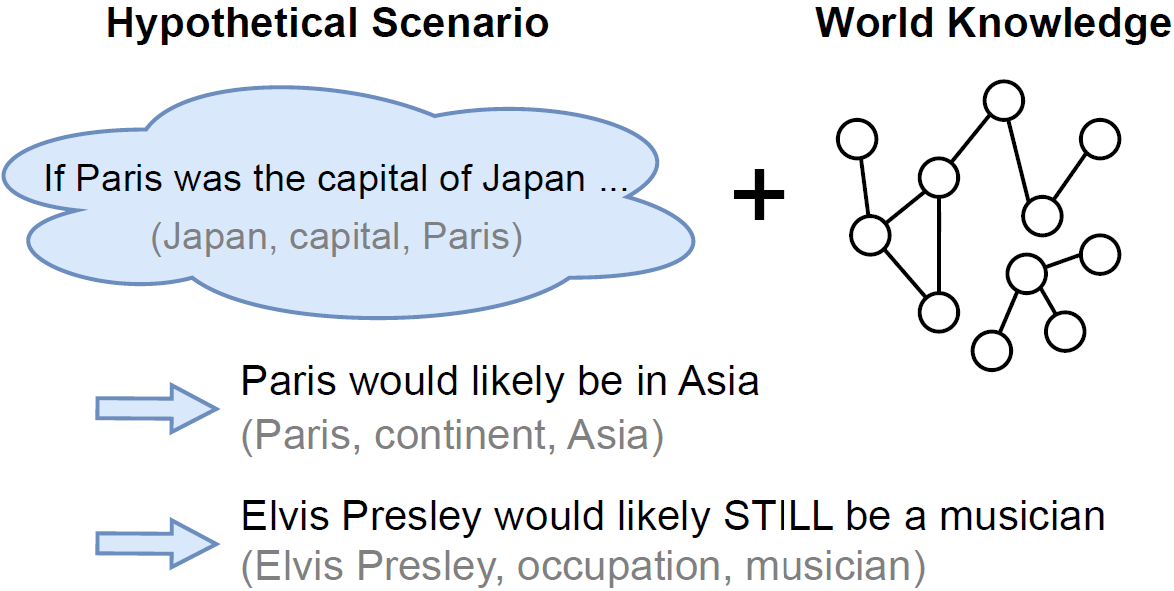}
\end{center}
\caption{A hypothetical scenario and its implications, expressed in the language of knowledge graph triples}\label{fig:fig_1}
\end{figure}

In this work, we link counterfactual reasoning to knowledge graph completion (KGC) via our new task \emph{\taskname}\footnote{The data and code are available at \url{https://github.com/LenaZellinger/counterfactual_KGR}.} (\textbf{C}ounter\textbf{F}actual \textbf{KG} \textbf{R}easoning) which requires models to classify the validity of facts given a hypothetical scenario. 
\taskname{} describes the original world state as a KG and hypothetical scenarios as edges that are added to the graph.
The hypothetical scenario leads to the emergence of new facts in the KG while leaving (most) already existing ones intact. 
Figure \ref{fig:fig_1} illustrates a hypothetical scenario in which Paris is the capital of Japan. 
To perform well on \taskname, models must be capable of detecting plausible additions to the graph, e.g., (Paris, continent, Asia), while maintaining knowledge of unaffected facts, e.g., (Elvis Presley, occupation, musician). We create the first benchmark datasets for \taskname, which are based on the CoDEx KGC benchmark \cite{safavi-koutra-2020-codex} and provide diverse hypothetical scenarios with corresponding plausible additions to the KG derived from \emph{inference rules} (that were mined from the KG \cite{amie3}).
We validate our data-generating process and underlying assumptions via thorough human annotation. Lastly, we introduce \emph{COULDD} (\textbf{COU}nterfactual Reasoning with Know\textbf{L}edge Graph Embe\textbf{DD}ings), a method which
updates existing KGEs based on counterfactual information. 
COULDD follows a standard KGE training scheme using the hypothetical scenario and negative sampling. Training stops once the hypothetical scenario is classified as valid.

In our experiments, COULDD is initialized with five different KGE methods. We observe that it can detect plausible counterfactual changes to the graph that follow prominent inference patterns in the KG while maintaining performance on unaffected triples. We repeat the same experiments with ChatGPT, i.e., gpt-3.5-turbo, provided with similar prompts to the human annotators.
ChatGPT performs better at detecting plausible additions to the graph than most KGE-based methods but exhibits poor knowledge retention. Qualitative analysis of answers provided by ChatGPT shows that it largely failed to understand the task on retained facts as it tried to infer them from the provided information. Evaluating on human-annotated data leads to a drop in overall performance for KGEs and ChatGPT alike. To summarize, our main contributions are as follows:
\begin{itemize}
    \item We propose \taskname, a challenging task for counterfactual reasoning on KGs and create corresponding, partially human-verified, datasets, which we make publicly available.
    \item We introduce COULDD, a general method for adapting existing KGE methods to make inferences given hypothetical scenarios and show that it improves reasoning on counterfactual graphs over pre-trained embeddings.
    \item We compare counterfactual reasoning with KGEs to ChatGPT and show that ChatGPT outperforms KGEs in detecting plausible counterfactual inferences but struggles to recall unrelated knowledge, unlike COULDD.
\end{itemize}
\section{\taskname: Task Description}
\label{sec:task}

We introduce \emph{Counterfactual KG Reasoning} (\emph{\taskname}{}) a novel task to assess the ability of machine learning systems to reason in hypothetical scenarios. 
\taskname{} describes the originally observed world state as a knowledge graph and introduces hypothetical scenarios by adding previously unseen facts to the graph. To perform well on \taskname, models need to (1) identify plausible changes to the original world state induced by the hypothetical scenario and (2) understand which facts are unaffected by the hypothetical scenario. 

\subsection{Definition of Counterfactual Graphs}
Formally, \taskname{} defines the original world state via a knowledge graph $\mathcal{G} = \{\mathcal{E}, \mathcal{R}, \mathcal{F}\}$, where $\mathcal{E}$ and $\mathcal{R}$ denote the sets of entities and relations represented in the knowledge graph.
The fact set $\mathcal{F}$ represents our knowledge about the world as triples $(h, r, t) \in \mathcal{F} \subset \mathcal{E} \times \mathcal{R} \times \mathcal{E}$. The fact set is usually split into disjoint subsets $\mathcal{F}_{train}$, $\mathcal{F}_{valid}$ and $\mathcal{F}_{test}$.
We denote a hypothetical scenario by a triple $\tau^c :=(h, r, t) \notin \mathcal{F}$. 
The \emph{counterfactual graph}, in which $\tau^c$ holds, is then characterized by the fact set $\mathcal{F}^c := \mathcal{F}\setminus \mathcal{F}^- \cup \mathcal{F}^+$, where
$\mathcal{F}^+$ denotes the facts that emerge given the hypothetical scenario,
and $\mathcal{F}^-$ denotes facts that contradict the scenario and cannot hold any longer. 
We say $\tau^c$ \textit{changes} a triple $\tau$ if either $\tau \in \mathcal{F}^+$ or $\tau \in \mathcal{F}^-$. 

In the following, we formulate the assumptions underlying our task.\\
\noindent
\textbf{Closed-world assumption.} We adopt the standard closed-world assumption \cite{Reiter1978}, which states that facts that are not part of the KG, i.e., $\tau \notin \mathcal{F}$, are false. \emph{Thus, each $\tau \notin \mathcal{F}$ is a possible hypothetical scenario in our setup}.

\noindent
\textbf{Logic-world assumption.}
We assume that plausible changes to the graph largely follow some regularity and can hence be modeled via (potentially very complex) logical rules.
While available rule sets have limited coverage and precision, we can leverage them to model a subset of plausible changes to a KG.
\emph{By employing the logic-world assumption, we can represent an approximation of $\mathcal{F}^c$ via a set of rules and the original fact set.}

\begin{figure*}[t]
    \begin{minipage}{0.48\textwidth}
        \centering
        \includegraphics[scale=0.45]{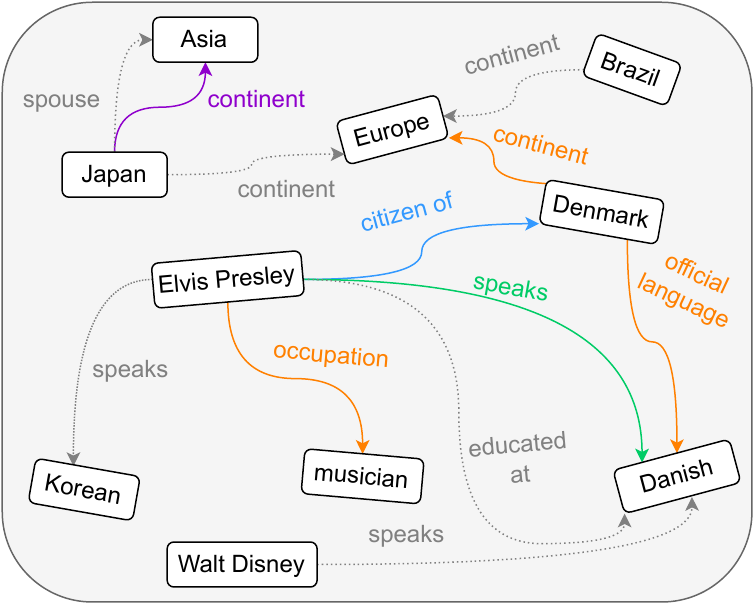}
    \end{minipage}
    \hfill
    \begin{minipage}{0.52\textwidth}
        \centering
        \resizebox{\textwidth}{!}{
        \begin{tabular}{l|l|l|l}\toprule
        \textbf{Instance} & \textbf{Notation} & \textbf{Original KG} & \textbf{CF KG}\\ \hline
    \textcolor{cyan}{Counterfactual} & $\tau^c$ & $\tau^c \notin \mathcal{F}$ & $\tau^c \in \mathcal{F}^c$\\\hline
     \textcolor{green}{Inference} & $\tau^i$ & $\tau^i \notin \mathcal{F}$ & $\tau^i \in \mathcal{F}^c$\\ \hline
     \textcolor{orange}{Unchanged (near)} & $\tau^n$ & $\tau^n \in \mathcal{F}$ & $\tau^n \in \mathcal{F}^c$\\\hline
     \textcolor{violet}{Unchanged (far)} & $\tau^f$ & $\tau^f \in \mathcal{F}$ & $\tau^f \in \mathcal{F}^c$\\\hline
    \textcolor{gray}{Corruptions} & $\tau_{h'}$, $\tau_{t'}$, $\tau_{r'}$ & $\tau_{h'}$, $\tau_{t'}$, $\tau_{r'} \notin \mathcal{F}$ & $\tau_{h'}, \tau_{t'}, \tau_{r'} \notin \mathcal{F}^{c}$\\\bottomrule
     \end{tabular}
     }
    \end{minipage}
    \vspace{1em}
    \caption{Overview over the different types of facts, given the hypothetical scenario that Elvis Presley is a citizen of Denmark. The green edge (Elvis Presley, speaks, Danish) emerges from adding the blue edge (Elvis Presley, citizen of, Denmark) to the knowledge graph. Purple and orange edges are present in the original KG and unaffected by the scenario. Grey edges are neither present in the original nor the counterfactual knowledge graph.}
    \label{fig:couldd_figure}
\end{figure*}

\subsection{Evaluation}\label{sec:evaluation}
We formulate \taskname{} as a binary classification task in which the goal is to predict whether a given triple is present in the counterfactual graph or not. 
Triples $ \tau \in \mathcal{F}^c$ receive label 1, while all other triples are labeled 0.
Since scoring all possible triples is infeasible, we consider a smaller set of carefully chosen test cases. Given a counterfactual $\tau^c \notin \mathcal{F}$ and a rule, we define:\\
(1) a \textbf{counterfactual inference $\tau^i$} that follows from the rule and allows us to measure whether the model can correctly predict changes to the graph given $\tau^c$,\\
(2) \textbf{retained facts} which are unaffected by the hypothetical scenario and should still be classified as valid in the counterfactual graph,\\
(3) random \textbf{head, tail, and relation corruptions} of inferences and retained facts, which ensure that the model does not predict unsolicited triples as valid additions. We denote the corruptions for a triple $\tau$ by $\tau_{h'}$, $\tau_{t'}$ and $\tau_{r'}$.\\
For (2), we distinguish between \textbf{near facts} $\tau^n$, which are in the one-hop neighborhood of $\tau^c$, and \textbf{far facts} $\tau^f$, sampled from its complement. Note that they are sampled from the entire fact set $\mathcal{F}$ to measure knowledge retention.
Figure \ref{fig:couldd_figure} illustrates a counterfactual scenario and its associated test cases. 

We use the following metrics to evaluate the performance on our benchmark. The concrete formulas can be found in Appendix \ref{sec:evaluation_metrics}. We compute\\
(1)  the \textbf{F1-score} over all test cases in the dataset to measure the overall predictive performance on counterfactual graphs. \\
\noindent
(2) the \textbf{accuracy on changed facts}, i.e., triples that have a different label before and after the hypothetical scenario is introduced.\\
\noindent
(3) the \textbf{F1-score on unchanged facts}, i.e., triples that have the same label before and after the hypothetical scenario is introduced.
\section{\taskname: Dataset Creation}\label{sec:cckg}
For our dataset construction, we leverage rules found by rule mining systems, which capture prominent patterns in KGs. Automatically mined rules are naturally compatible with the content of the KG and are known to be a useful tool for KGC \cite[e.g.,][]{meilicke2019anytime, drum}. Since there is no trivial way to \emph{reliably} generate $\mathcal{F}^-$, we only consider the additions $\mathcal{F}^+$. Concretely, we define $\mathcal{F}^+$ via mined \emph{composition rules} of the form
\begin{equation}\label{eq:comp_rule}
(X, r_1, Y) \land (Y, r_2, Z) \rightarrow (X, r_3, Z)
\end{equation}
where $r_1, r_2, r_3 \in \mathcal{R}$. We refer to $(X, r_1, Y) \land (Y, r_2, Z)$ as the \emph{rule body} and $(X, r_3, Z)$ as the \emph{inference}. The triples $(X, r_1, Y)$ and $(Y, r_2, Z)$ are called the first and second body atom, respectively. Replacing $X$, $Y$, and $Z$ by concrete entities $x,y,z \in \mathcal{E}$ creates an \emph{instantiation} of the rule. In the following, we will use the short-hand notation $(r_1, r_2, r_3)$ to denote a rule as described in (\ref{eq:comp_rule}).

We choose composition rules since they are well studied in standard KG completion benchmarks \cite{safavi-koutra-2020-codex} and inferential benchmarks \cite{cao-etal-2021-missing, liu2023revisiting}. Moreover, composition rules, as given in (\ref{eq:comp_rule}), infer \emph{local changes}. This is desirable since most relevant changes induced by a hypothetical scenario will likely occur in its close neighborhood. We consider understanding the implications induced by composition rules as a first step to more general and complex hypothetical reasoning.

\subsection{Data Generating Process}\label{sec:data_generating_process}

In the following, we give a high-level overview of our data generating process. We focus on creating hypothetical scenarios for the first body atom of a given rule. Appendix \ref{sec:cckg_appendix} provides a detailed description and the full algorithm.

\begin{figure}[t!]
\centering
\includegraphics[scale=0.9]{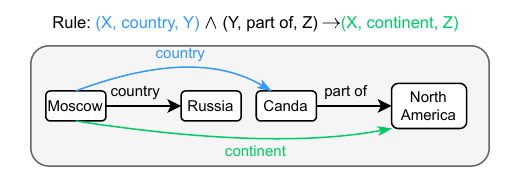}
\caption{Creation of a hypothetical scenario.}\label{fig:cckg}
\end{figure}

Given a knowledge graph and a rule set, we generate several hypothetical scenarios for each rule by altering a fact in the KG such that it triggers the rule, as is illustrated in Figure \ref{fig:cckg}. Concretely, for each rule $(r_1, r_2, r_3)$, we search for existing edges $e_1 := (x, r_1, y) \in \mathcal{F}_{train}$ and $e_2 := (\bar{y}, r_2, z) \in \mathcal{F}_{train}$, ensuring that the resulting hypothetical scenario $\tau^c:=(x, r_1, \bar{y})$ and inference $\tau^i := (x, r_3, z)$ are not in the original KG. Sampling $e_1$ and $e_2$ without any constraints can result in nonsensical scenarios and inferences. Hence, we ensure that the entities in $\tau^c$ and $\tau^i$ are suitable for the given relation by restricting them to entities that occur with said relation in the original KG. Once suitable $\tau^c$ and $\tau^i$ are found, we randomly sample two near facts $\tau^n$ from the one-hop neighborhood of $\tau^c$ and one far fact $\tau^f$ from its complement. Note that we sample $\tau^n$ and $\tau^f$ on the full fact set $\mathcal{F}$, instead of only $\mathcal{F}_{test}$, as their primary purpose is to measure knowledge retention as opposed to inference capabilities. 

When creating head and tail corruptions of a given fact, we restrict the sample space since random corruptions, which tend to result in nonsensical triples, have previously been shown to be easily detectable for KGE methods \cite{safavi-koutra-2020-codex}. For head (tail) corruptions, we require that the replacements are also heads (tails) for the relation in the original graph\footnote{In rare cases where these constraints only allow for creating triples already present in the KG or inferred by our rule set, we default to the full entity set.}. For relation corruptions, we do not employ additional constraints.

\subsection{\taskname-CoDEx}
Based on the procedure described in Section \ref{sec:data_generating_process}, we create the first benchmark datasets for \taskname{} based on the CoDEx knowledge graph completion benchmark \cite{safavi-koutra-2020-codex}. We choose CoDEx since it covers diverse content, uses easily interpretable relations, and contains rich auxiliary information, such as entity types.
CoDEx provides three knowledge graphs of varying sizes (S, M, and L), collected from Wikidata \cite{wikidata}, and corresponding \emph{composition rules obtained by the rule-mining system Amie3} \cite{amie3}. CoDEx-S and CoDEx-M additionally contain verified negative triples. An overview over the resources provided by CoDEx can be found in Appendix \ref{sec:codex_resources}.

We use the available Amie3 patterns for each CoDEx dataset as our rule set and create at most 25 unique counterfactual triples per body atom for each rule. We subsequently split them into a validation and test set, ensuring that there are no overlapping rules or counterfactuals between validation and test \footnote{For M, there are rules which can produce the same counterfactual - inference pairs (using a different context). There are 14 such duplicates in the test set. Still, there is no overlap in counterfactuals between validation and test.}. Table \ref{tab:cckg_stats} provides statistics about the created datasets.
 
In the following section, we will explore how well the resulting test cases align with \emph{human counterfactual reasoning}.

\begin{table}
\resizebox{0.48\textwidth}{!}{
\begin{tabular}{lcccc}\toprule
& \multicolumn{2}{c}{Valid} & \multicolumn{2}{c}{Test}\\
& Rules & Facts & Rules & Facts\\\midrule
\taskname-CoDEx-S & 5 & 3600 & 12 & 8848\\
\taskname-CoDEx-M & 5 & 3936 & 26 & 19584\\
\taskname-CoDEx-L & 5 & 4000 & 39 & 30064\\\bottomrule
\end{tabular}
}
\caption{\taskname{} dataset overview. "Rules" denotes the number of rules that were used to create the dataset. "Facts" is the total number of test cases. }\label{tab:cckg_stats}
\end{table}

\subsection{Human Annotation}
\label{sec:verifacation}

We validate our data generating process via human annotation. For each of the 31 rules in \taskname-M, we verify 10 test instances (5 per atom\footnote{Except for one rule which only produced one unique counterfactual according to our conditions for the second atom.}). We annotate $\tau^i$, $\tau^f$, $\tau^n_1$, $\tau^n_2$ and $\tau^i_{r'}$, and omit the remaining corruptions as their construction relies on the commonly-used closed-world assumption \cite{Reiter1978}. This results in 1530 annotated instances, which were labeled by four to six annotators as either \emph{likely} (1), \emph{unlikely} (0), or \emph{unsure/too little information} (-1), given verbalizations of the hypothetical scenario and context triggering the respective inference rule. We observe a Krippendorff's alpha \cite{hayes2007answering} of 0.653, computed using the simpledorff library, which indicates substantial agreement \cite{landis1977measurement}.
The annotation guidelines can be found in Appendix \ref{sec:human_verification_appendix}. Table \ref{tab:annotation_results} summarizes the annotation results.

Inferences seem to be the most difficult category to annotate as they show the highest amount of ties and "unsure/too little information" labels. Moreover, we observe the highest number of deviations from our expected label for this test case.
This indicates that rules that were mined for \emph{factual} knowledge graph completion cannot always be used for human-like counterfactual reasoning.

On relation corruptions, we observe a noticeable number of inferences that are not implied by our rules, but are still considered valid by humans or are at least debatable. Possible explanations are the limited coverage of the rule set or unintuitive verbalizations of the relations. For near and far facts, we obtain a label distribution that largely agrees with our assumptions.
\begin{table}[t!]
    \centering
    \resizebox{0.47\textwidth}{!}{
    \begin{tabular}{lccc|cccc}\toprule
    & & &  & \multicolumn{4}{c}{Majority Vote Label}\\
     & \# Labeled & Expected & As expected & 0 & 1 & -1 & Tied \\
    \midrule
    Inference & 306 & 1 & 58.2\% & 60 & 178 & 27 & 41 \\
    Far fact & 306 & 1 & 99.7\% & 0 & 305 & 0 & 1 \\
    Near fact & 612 & 1 & 95.6\% & 16 & 585 & 2 & 9 \\
    Relation corr. & 306 & 0 & 86.9\% & 266 & 20 & 3 & 17 \\\bottomrule
    \end{tabular}
    }
    \caption{Annotation results. "\# Labeled" denotes the number of annotated examples per category. "Expected" gives the label assigned by our automatic process and "As expected" gives the percentage of samples for which the expected label coincides with the majority vote.}\label{tab:annotation_results}
\end{table}

\section{Counterfactual Reasoning with Knowledge Graph Embeddings}\label{sec:kge_methods}
\begin{table*}[ht!]
\centering
\resizebox{0.95\textwidth}{!}{
\begin{tabular}{llllllllll}
\toprule
 & \multicolumn{3}{c}{CFKGR-CoDEx-S} & \multicolumn{3}{c}{CFKGR-CoDEx-M} & \multicolumn{3}{c}{CFKGR-CoDEx-L} \\
 & F1 & Changed & Unchanged & F1 & Changed & Unchanged & F1 & Changed & Unchanged \\
\midrule
RESCAL & 60.82 & 27.12 & 63.28 & 63.05 & 21.57 & 66.92 & 53.84 & 71.47 & \textbf{49.64} \\
COULDD-RESCAL & \textbf{61.68 $\pm$ 0.14} & \textbf{32.48 $\pm$ 0.73} & \textbf{63.48 $\pm$ 0.16} & \textbf{63.85 $\pm$ 0.08} & \textbf{26.23 $\pm$ 0.16} & \textbf{67.16 $\pm$ 0.07} & \textbf{53.94 $\pm$ 0.02} & \textbf{84.56 $\pm$ 0.35} & 48.18 $\pm$ 0.06 \\ \midrule
TransE & 58.94 & 23.15 & 61.87 & 53.61 & 23.61 & \textbf{55.83} & 49.23 & 66.31 & 45.37 \\
COULDD-TransE & \textbf{60.49 $\pm$ 0.12} & \textbf{26.8 $\pm$ 0.81} & \textbf{63.16 $\pm$ 0.09} & \textbf{53.91 $\pm$ 0.05} & \textbf{26.06 $\pm$ 0.25} & 55.79 $\pm$ 0.06 & \textbf{52.6 $\pm$ 0.06} & \textbf{76.56 $\pm$ 0.25} & \textbf{47.77 $\pm$ 0.04} \\ \midrule
ComplEx & 62.45 & 29.11 & 64.90 & 65.69 & 11.60 & \underline{\textbf{71.83}} & 58.44 & 65.51 & \textbf{55.26} \\
COULDD-ComplEx & \underline{\textbf{67.76 $\pm$ 0.3}} & \textbf{37.94 $\pm$ 0.67} & \underline{\textbf{69.95 $\pm$ 0.29}} & \underline{\textbf{66.78 $\pm$ 0.06}} & \textbf{34.67 $\pm$ 0.23} & 69.21 $\pm$ 0.07 & \textbf{59.44 $\pm$ 0.02} & \textbf{82.95 $\pm$ 0.26} & 54.25 $\pm$ 0.02 \\ \midrule
ConvE & 61.04 & 16.64 & 65.39 & \textbf{56.83} & 13.15 & \textbf{61.37} & 55.56 & \textbf{61.84} & 52.58\\
COULDD-ConvE & \textbf{61.51 $\pm$ 0.11} & \textbf{16.96 $\pm$ 0.72} & \textbf{65.92 $\pm$ 0.12} & 52.69 $\pm$ 0.16 & \textbf{17.04 $\pm$ 0.16} & 56.09 $\pm$ 0.16 & \underline{\textbf{60.6 $\pm$ 0.17}} & 45.53 $\pm$ 0.61 & \underline{\textbf{60.29 $\pm$ 0.14}} \\ \midrule
TuckER & 64.25 & 15.01 & \textbf{69.40} & 65.21 & 13.15 & \textbf{70.98} & 52.87 & 76.74 & \textbf{48.05} \\
COULDD-TuckER & \textbf{66.03 $\pm$ 0.13} & \textbf{35.99 $\pm$ 1.0} & 68.09 $\pm$ 0.19 & \textbf{66.09 $\pm$ 0.17} & \textbf{43.69 $\pm$ 0.38} & 66.95 $\pm$ 0.17 & \textbf{53.53 $\pm$ 0.04} & \underline{\textbf{88.47 $\pm$ 0.34}} & 47.49 $\pm$ 0.02 \\ \midrule
gpt-3.5-turbo &
47.83 & \underline{68.90} & 40.22 &
46.72 & \underline{52.12} & 42.25 &
45.80 & 52.10 & 40.95\\
\bottomrule
\end{tabular}
}
\caption{Test performance of pre-trained embeddings and COULDD on \taskname. For COULDD, we report the mean and standard deviation across 5 runs. Bold entries denote the best performance between pre-trained KGEs and their counterpart trained with COULDD. The best results on the dataset are underlined. For all scores, higher is better.}\label{tab:main_table}
\end{table*}

KGE models find low-dimensional vector representations for entities and relations while preserving the information contained in the KG. 
To judge the plausibility of a given triple, KGE models use a scoring function $\phi(h, r, t): \mathcal{E} \times \mathcal{R} \times \mathcal{E} \rightarrow \mathbb{R}$. A triple is typically classified as valid if it satisfies $\phi(h, r, t) \geq \mu_r$, for a relation-specific threshold $\mu_r \in \mathbb{R}$.

 \begin{algorithm}
\caption{COULDD training and prediction. The short-hand notation $\phi_{\theta}(\mathcal{T}_{\tau^c})$ denotes scoring all test cases associated with $\tau^c$ and $\mathcal{L}_{\theta}$ denotes the cross-entropy loss.}\label{alg:couldd}
\DontPrintSemicolon
\KwData{$\mathcal{G} = \{\mathcal{E}, \mathcal{R}, \mathcal{F}\}$,\newline
\taskname{} data $\mathcal{D}$,\newline
original embeddings $\theta_0$,\newline
\# iterations $E$,\newline
\# additional samples $N$,\newline
learning rate $\alpha$,\newline
thresholds $\mu_1, \mu_2, ..., \mu_{|\mathcal{R}|}$
}
\KwResult{\taskname{} predictions}
$\hat{y} \leftarrow$ \{\}\\
\ForEach{$(\tau^c, \mathcal{T}_{\tau^c}) \in \mathcal{D}$}{
$\theta \leftarrow \theta_0$\; 
\For{$e \in \{1, ..., E\}$}{
$S \leftarrow $ Sample $N$ from $\mathcal{F}_{train}$\;
$B \leftarrow \{\tau^c \} \cup S$\;
$\theta \leftarrow \text{Optimizer}(\mathcal{L}_{\theta}(B), \alpha)$\;
\If{$\phi_{\theta}(\tau^c) \geq \mu_r$}{
break
}
}
$\hat{y} \leftarrow \hat{y} \cup \{\phi_{\theta}(\mathcal{T}_{\tau^c})\}$
}
\Return{$\hat{y}$}
\end{algorithm} 
To extend KGEs to our task, we propose \emph{COULDD} (\textbf{COU}nterfactual Reasoning With Know\textbf{L}edge Graph Embe\textbf{DD}ings), a general method for adapting existing knowledge graph embeddings with respect to a given hypothetical scenario. COULDD is initialized from existing embeddings trained on the original KG. For each hypothetical scenario, these embeddings are updated and subsequently evaluated on the corresponding test cases.

COULDD's update scheme only minimally changes standard KGE training: In each iteration, the existing embeddings are fine-tuned on a batch consisting of the counterfactual triple $\tau^c$ and $N$ additional randomly sampled edges from the training graph. Negative training examples are generated by randomly corrupting the head and tail entities of each triple in the batch. The embeddings are updated using the standard cross-entropy loss. Once the counterfactual triple $\tau^c$ exceeds the classification threshold, the training is stopped in order to avoid an excessive perturbation of the pre-trained embeddings\footnote{Note that there is no traditional validation set for the individual updates on which we could perform early stopping.}. 

Importantly, COULDD only requires access to the counterfactual triple $\tau^c$ and the original fact set $\mathcal{F}$ and does not require additional task-specific training data or information about the rules used to generate \taskname{} datasets\footnote{We only use the test cases in the validation set for hyperparameter tuning.}. As a result, COULDD can also be applied in rule-free evaluation setups. Algorithm \ref{alg:couldd} provides a formal description of COULDD.

\section{Experiments}
\label{sec:experiments}
\begin{table*}[t]
\begin{center}
\resizebox{0.95\textwidth}{!}{
\begin{tabular}{lllllll|ll}
\toprule
& \multicolumn{6}{c|}{\taskname{}-CoDEx-M*} &  \multicolumn{2}{c}{CoDEx-M (filtered)}\\
 & F1 (E) & F1 (H) & Changed (E) & Changed (H) & Unchanged (E) & Unchanged (H) & Overall & Rule-wise \\
\midrule
RESCAL &
\textbf{89.30} & \textbf{87.61} &
21.55 & 13.64 &
\textbf{97.20} & \textbf{96.17} &
92.74 & 84.72\\
COULDD-RESCAL & 
89.03 $\pm$ 0.24 & 87.12 $\pm$ 0.24 &
\textbf{25.08 $\pm$ 0.75} & \textbf{16.25 $\pm$ 0.58} &
96.48 $\pm$ 0.20 & 95.31 $\pm$ 0.21 &
$-$ & $-$\\\hline
TransE &
\textbf{81.21} & \textbf{79.85} &
21.55 & 16.48 &
\textbf{88.55} & \textbf{87.73} &
91.29 & 80.26\\
COULDD-TransE &
80.64 $\pm$ 0.07 & 79.44 $\pm$ 0.10 &
\textbf{23.43 $\pm$ 0.27} & \textbf{19.2 $\pm$ 0.43} &
87.65 $\pm$ 0.11 & 86.94 $\pm$ 0.12 &
$-$ & $-$\\\hline
ComplEx &
89.01 & 87.53 &
9.94 & 2.84 &
\textbf{98.40} & \underline{\textbf{97.51}} &
96.01 & 77.79\\
COULDD-ComplEx &
\textbf{92.05 $\pm$ 0.11} & \textbf{90.43 $\pm$ 0.16} &
\textbf{37.35 $\pm$ 1.08} & \textbf{29.89 $\pm$ 1.37} &
98.29 $\pm$ 0.1 & 97.27 $\pm$ 0.1 &
$-$ & $-$\\\hline
ConvE & 
\textbf{83.96} & \textbf{82.56} &
14.92 & 9.09 &
\textbf{92.46} & \textbf{91.62} &
89.29 & 79.70\\
COULDD-ConvE &
78.39 $\pm$ 0.56 & 77.15 $\pm$ 0.72 &
\textbf{16.69 $\pm$ 1.13} & \textbf{12.39 $\pm$ 0.91} &
86.17 $\pm$ 0.62 & 85.43 $\pm$ 0.71 &
$-$ & $-$\\\hline
TuckER &
89.31 & 88.08 &
13.81 & 7.95 &
98.26 & \textbf{97.50} &
\underline{96.37} & \underline{90.33}\\
COULDD-TuckER & 
\underline{\textbf{92.83 $\pm$ 0.12}} & \underline{\textbf{90.92 $\pm$ 0.12}} &
\textbf{43.43 $\pm$ 0.90} & \textbf{34.55 $\pm$ 0.91} &
\underline{\textbf{98.41 $\pm$ 0.11}} & 97.21 $\pm$ 0.12 &
$-$ & $-$\\\hline
gpt-3.5-turbo & 
63.96 & 63.36 & \underline{53.04} & \underline{53.98} & 62.75 & 62.34 &
$-$ & $-$\\
\bottomrule
\end{tabular}
}
\caption{Case study on \taskname{}-CoDEx-M* with expected (E) and human-assigned (H) labels and performance on the filtered CoDEx-M test set. "Overall" describes the accuracy across all inferences. "Rule-wise" gives the average accuracy per rule. Bold entries denote the best performance between pre-trained KGEs and their counterpart trained with COULDD. The best results on the dataset are underlined. For all scores, higher is better.}\label{tab:case_study}
\end{center}
\end{table*}

In the following, we conduct two types of experiments:
First, we evaluate pre-trained KGEs, COULDD, and ChatGPT on our \taskname{} datasets with expected labels to assess whether the methods can apply inference rules found by a rule mining system in hypothetical scenarios. In our second set of experiments, we evaluate on human-labeled data to check whether the methods also capture human reasoning, which does not necessarily align with mined inference rules (see Section \ref{sec:verifacation}).

\subsection{General Setup}

We use the five pre-trained CoDEx link-prediction models as initializations for COULDD\footnote{The config files for the models are available at \url{https://github.com/tsafavi/codex}}.
Further details about the KGE methods are in Appendix \ref{sec:architectures}.

For COULDD, we tune the learning rate ($\alpha$) and number of additional samples per batch ($N$) on the respective \taskname{} validation set, based on the best overall F1-score, and set the maximum number of update steps ($E$) to 20. We carry over the remaining hyperparameters from the pre-trained CoDEx models \cite{safavi-koutra-2020-codex}. Further details regarding the hyperparameters are in Appendix \ref{sec:hyperparameters_app}.
Optimization is performed using Adam \cite{kingma2014adam}, or Adagrad \cite{duchi2011adagrad}, depending on the original model configuration. 
The general classification setup and relation-specific decision thresholds are equivalent to the original CoDEx paper\footnote{We added a minor correction to the CoDEx threshold tuning that ensures proper application of the global threshold for unobserved relations.} \citep{safavi-koutra-2020-codex} to ensure comparability. Note that this entails scoring all triples in the tail direction. Since no negatives are provided for CoDEx-L, we generate one random tail corruption per validation triple for threshold tuning (akin to experiments in \cite{safavi-koutra-2020-codex}).
During training, we sample 100 negative examples per triple (50 head and 50 tail corruptions), as this was effective in previous work \cite{pmlr-v48-trouillon16, kotnis2017analysis}.

We implement our experiments by adapting LibKGE \cite{libkge} to support our proposed COULDD training strategy. We perform hyperparamter optimization using Optuna \cite{optuna}. For experiments with ChatGPT, i.e., \textit{gpt-3.5-turbo}, we use the OpenAI API and temperature 0. The used prompts and an example of input and output can be found in Appendix \ref{sec:appendix_gpt}.

\subsection{Results}
Table \ref{tab:main_table} contains the results. A detailed evaluation per test type can be found in Appendix \ref{sec:eval_per_test_type}.
First, we observe that the KGE performances on \taskname{}-CoDEx-L differ noticeably from \taskname{}-CoDEx-S and \taskname{}-CoDEx-M. This is likely due to lower threshold quality resulting from the absence of hard negative triples for CoDEx-L.

COULDD achieves the best results in terms of overall F1-score on all datasets. In particular, COULDD noticeably improves the performance on changed facts over the pre-trained embeddings, except for ConvE. Importantly, we do not observe a case where applying COULDD leads to a noticeable loss of knowledge acquired during pre-training. In terms of overall F1-score, COULDD-ComplEx achieves the best results averaged across the three datasets. On changed facts, COULDD-TuckER is the best-performing KGE method, likely because TuckER is well-suited for modeling compositional relations \cite{safavi-koutra-2020-codex}. ChatGPT achieves the best scores on changed facts on two out of three datasets. However, it generally does not perform well on unchanged facts. Possible reasons are that it misses relevant background knowledge present in the KG or does not understand the task on these instances. 
In summary, we observe that COULDD consistently improves performance over the pre-trained embeddings, overall and on changed facts in particular, and does not strongly degrade performance on unchanged facts. This indicates that COULDD, to an extent, can be used to infer plausible counterfactual changes to the graph when they follow prominent patterns in the KG.

\subsection{Case Study on CoDEx-M}\label{sec:case_study}

To better understand the results shown in Table \ref{tab:main_table}, we conduct a case study on CoDEx-M for which we have a human-annotated \taskname{} subset. In particular, we want to assess how well the pre-trained CoDEx models perform \emph{factual} reasoning with composition rules and how an evaluation on human-assigned labels affects our results.
The main results are presented in Table \ref{tab:case_study}. Table \ref{tab:perf_analysis} in the appendix presents a confusion matrix per test type for COULDD and ChatGPT.

\subsubsection{Inference Rules in Factual Contexts}\label{sec:inferential_case_study}
Achieving good performance on changed triples in Table \ref{tab:main_table} requires (1) a logical adaption to the hypothetical scenario and (2) the application of composition rules that generated the test inferences. We attempt to disentangle these factors by investigating whether the CoDEx models captured the regularities expressed in the Amie3 rules during pre-training and can apply them in \emph{factual} scenarios.\\
\noindent
\textbf{Setup.}
We filter the original CoDEx-M test set for triples that can be inferred by applying the mined Amie3 rules to the training set. We only keep triples that stem from rules that cover at least five triples in the test set to obtain sensible estimates for the rule-wise performances. This results in a filtered test set of 551 instances inferred from 10 rules.

\noindent
\textbf{Results.}
We notice that the performance on the filtered CoDEx-M test set is consistently high for all pre-trained KGE methods. This indicates that they learned the mined inference patterns during training and lower performances observed in Table \ref{tab:main_table} are likely due to insufficient adaption to the hypothetical scenario. Appendix \ref{sec:inferential_appendix} gives a full overview of rule-wise performances.

\begin{table*}
\resizebox{0.95\textwidth}{!}{
\centering
\begin{tabular}{l|l}
\toprule
Hypothetical: & Joe Frazier (human) died in Moscow (city with millions of inhabitants, largest city, capital)\\\hline
Context: & Joe Frazier (human) was cast in Encino Man (film)\\\hline
Statement: & Is it then still plausible that E-40 (human) has the occupation actor (profession, artist),\\& given that this is the case in the real world?\\\midrule
Answer by ChatGPT: & unlikely\\\hline
Explanation by ChatGPT: & The hypothetical scenario states that Joe Frazier died in Moscow, which is a city with millions of inhabitants. \\& This fact is not the case in the real world. The context mentions that Joe Frazier was cast in the film Encino Man. \\& \textbf{Since the statement is 'Is it then STILL plausible that E-40 has the occupation actor?',} \\& \textbf{we need to consider if the hypothetical scenario makes it plausible for E-40 to have the occupation of an actor.} \\& \textbf{However, the context does not provide any specific information about E-40 or his occupation.} \\& Therefore, we cannot determine the plausibility of E-40 being an actor based on the given information.\\& Hence, the statement is unlikely.\\\bottomrule
\end{tabular}
}
\caption{Example prompt and corresponding answer produced by ChatGPT on \taskname{}-CoDEx-M*.}\label{tab:gpt_examples_main_text}
\end{table*}

\subsubsection{Evaluation with Manual Labels}\label{sec:kge_manual}

\textbf{Setup.} We create a subset of the \taskname{}-CoDEx-M test set, denoted by \textbf{\taskname{}-CoDEx-M*}, which solely consists of instances that were manually labeled as described in Section \ref{sec:verifacation}.
We filter out test cases for which less than three annotators agreed or the majority vote label was "unsure/too little information".
This results in a dataset composed of 256 hypothetical scenarios and a total of 1159 test cases.
The complete label distribution per test case can be found in Appendix \ref{sec:cckg_m_star_stats}.
\\
\textbf{Results.} 
First, we note that the overall F1-score and performance on unchanged facts greatly differ from the numbers observed for the unfiltered set in Table \ref{tab:main_table}. This is due to the omission of most corruptions as they were not manually labeled. We observe a consistent performance drop for KGE-based methods when evaluating on human-assigned labels instead of expected labels for all metrics. Judging from the confusion matrix in Table \ref{tab:perf_analysis}, COULDD cannot reliably identify \emph{false inferences} that follow from patterns in the KG, but are invalid according to the annotators. Moreover, it classifies most \emph{outdated facts}, which are no longer valid given the hypothetical scenario, as positive. However, the number of outdated facts (14) is too small to draw any substantial conclusions from this observation.\\
For ChatGPT, we observe a slightly reduced overall performance when evaluating with human-assigned labels. However, ChatGPT's score improves on changed facts for human-assigned labels. A closer look at the confusion matrix reveals that ChatGPT performs better at detecting outdated facts and false inferences than KGEs. However, as observed before, ChatGPT tends to misclassify facts that should be retained. A qualitative inspection reveals that ChatGPT largely misunderstands the task on such triples: instead of answering whether they STILL hold given the hypothetical scenario, it oftentimes tries to infer them. Table \ref{tab:gpt_examples_main_text} gives an example.
\section{Related Work}
\textbf{Inferential KGC Benchmarks.}
Rule-based \emph{inferential benchmarks} for KGC \cite{liu2023revisiting, cao-etal-2021-missing} assess a method's ability to learn implicit rule patterns and use them to predict inferences in the test set based on evidence in the training set. 
\citet{cao-etal-2021-missing} create an inferential test set for CoDEx-M based on a rule set mined by AnyBurl \cite{meilicke2019anytime}, akin to our experiments in Section \ref{sec:inferential_case_study}, and also find that pre-trained KGEs have strong inferential reasoning capabilities.
\\
\textbf{Counterfactual Graph Learning.}
Leveraging counterfactuals in graph learning is an emerging field of research \cite{guo2023counterfactual}. Counterfactuals have recently been utilized to ensure the fairness of graph-based systems with respect to sensitive node attributes \cite{agarwal2021towards, ma2022learning, zhang2021multi}, improve interpretability by generating counterfactual explanations for predictions \cite{lucic2022ter, numeroso2021meg, prado2022ensemble, xu2022counterfactual}, and enhance link prediction performance on the graph \emph{as-is} \cite{chang2023knowledge, lu2023causal, shi2022guided, wang2021incorporating, zhao2022counterfactual}.

Our work does not fall into any of the above categories and instead focuses on making predictions in a counterfactual graph.\\
\textbf{CF Reasoning Benchmarks for LLMs.}
Several datasets and evaluation schemes have been proposed for assessing the counterfactual reasoning capabilities of LLMs. 
\citet{qin-etal-2019-counterfactual} introduce the task of counterfactual story rewriting, in which LLMs have to minimally revise a given story with respect to a counterfactual event. 
The CRASS benchmark challenges LLMs to select a valid consequence given a questionized counterfactual conditional in a multiple-choice setting \cite{frohberg2022crass}.
\citet{li-etal-2023-counterfactual} present LLMs with a hypothetical premise and two possible completions for a corresponding statement, one of which is valid in the real world while the other holds in the hypothetical scenario.

In contrast, \taskname{} poses a binary classification task, in which the model has to decide whether a presented statement is plausible in the given hypothetical scenario or not. Further, our benchmark is based on the knowledge contained in a KG and thus considers specific, real-world entities.
\section{Discussion}\label{sec:discussion}
\textbf{Comparison with Human CF Reasoning.}
Our labeling efforts and experiments show that counterfactual reasoning on KGs is a challenging task. Both KGEs and ChatGPT leave much room for improvement on \taskname.
Moreover, as indicated by our annotation results (Table \ref{tab:annotation_results}), even humans find it difficult to judge the plausibility of KG-based counterfactual statements, especially when they involve unfamiliar situations. For instance, "If Meg White was a member of Girls Aloud, would Jack White be part of Girls Aloud?" is a question that most humans likely do not ask themselves.
Nevertheless, automatic systems can be presented with and evaluated on a wide range of possible scenarios, even if those are implausible or hard to imagine for humans.

\noindent
\textbf{Advantage of KG-based Benchmarks.}
KGs are a powerful tool for defining hypothetical scenarios and their consequences. 
The rich world knowledge stored in KGs allows to create interesting \emph{case-specific} inferences. In the example question above, would the judgement change if we replace "Girls Aloud" by a band that is not a girl group? This aspect is largely missing from current counterfactual reasoning benchmarks for LLMs \cite{frohberg2022crass, li-etal-2023-counterfactual}, as they mostly handle generic entities.

\section{Conclusion}
This work introduces the novel task \taskname{}, which requires models to reason on a counterfactual KG. By utilizing the world knowledge stored in KGs, we create datasets consisting of diverse hypothetical scenarios and their implications, as defined by inference rules. Further, we propose COULDD, a general method for counterfactual reasoning on KGs, and evaluate its effectiveness on automatically generated and human-annotated data. We extend our experiments to ChatGPT and find that it generally outperforms COULDD at making counterfactual inferences. However, ChatGPT largely does not recognize which facts are invariant to the hypothetical scenario. Both COULDD and ChatGPT leave much headroom on the task, highlighting the difficulty of \taskname{}.
\section{Limitations}
The type of rules that we examine is arguably limited. We consider understanding the implications induced by composition rules as a first step to more general and complex hypothetical reasoning. Moreover, while the set of outdated facts $\mathcal{F}^-$ is a key component for defining the counterfactual KG, there is no trivial way for generating them reliably without appropriate rules or extensive human verification. Most rules defined for KGs are Horn rules \cite[e.g.,][]{amie3, meilicke2019anytime, sadeghian2019drum} and do not express negation in the head atom. Hence, we focus on the additions $\mathcal{F}^+$ in this work.

Furthermore, this work does not consider the confidences of the mined Amie3 rules but assumes that they all could be a valid inference rules for hypothetical reasoning. As indicated by our human annotation results, this is likely not true in practice.

Verbalizing KG triples, in a way that is intuitive to humans, is a difficult task. We tried our best to find suitable verbalizations for the relations in the CoDEx KG by consulting the corresponding Wikidata definitions as well as ParaRel \cite{elazar-etal-2021-measuring}. In our verbalizations, each entity is presented with up to three of its associated entity types\footnote{Whenever more than three entity types were available, we randomly sampled three of them to enhance readability.} in order to facilitate reasoning with lesser-known entities.
Nevertheless, unintuitve verbalizations and missing context from the KG (with respect to how relations are used) might have influenced our annotation results and ChatGPT experiments.

Moreover, KGs can contain erroneous or outdated facts and automatically constructed \taskname{} examples might rely on these facts. It is possible that such instances impacted the performance of ChatGPT on our benchmark.

Lastly, the poor performance of ChatGPT on unchanged facts could partially be caused by the system prompt used in our experiments, which can be found in Appendix \ref{sec:appendix_gpt}. We designed the prompt based on the instructions provided to the human annotators. Nevertheless, it is likely that the prompt could be adjusted to improve the results of ChatGPT on unchanged facts. Appendix \ref{sec:chatgpt_observations} further details some frequent errors we noticed in ChatGPT's responses.

\section{Ethics Statement}
We relied on well-established and publicly available resources to build our datasets and method. We use the CoDEx knowledge graph and LibKGE, which are both published under the MIT license. The config files for the pre-trained CoDEx models used in our experiments are available on the CoDEx github repository\footnote{\url{https://github.com/tsafavi/codex}}.

The counterfactual situations included in our datasets are randomly generated and purely hypothetical. They do not convey any implications about the real-world entities referenced in them. Nevertheless, the created instances could be biased towards certain entities due to biases in the original KGs and our employed sampling strategy detailed in Appendix \ref{sec:cckg_appendix}.

We recruited annotators on a voluntary basis. We do not publish any information that could be used to identify the labelers and our data does not contain any personal information regarding the annotators.
\section*{Acknowledgements}
This research has been funded by the Deutsche Forschungsgemeinschaft (DFG, German Research Foundation) RO 5127/2-1 and the Vienna Science and Technology Fund (WWTF)[10.47379/VRG19008] ``Knowledge-infused Deep Learning for Natural Language Processing''.
We thank the European High Performance Computing initiative for providing the computational resources that enabled this work. EHPC-DEV-2022D10-051.

\bibliography{anthology,custom}
\bibliographystyle{acl_natbib}

\appendix

\label{sec:appendix}
\section{Evaluation Metrics}\label{sec:evaluation_metrics}
This section gives the concrete formulas for the metrics used in Table \ref{tab:main_table} and Table \ref{tab:case_study}. We denote the full evaluation dataset by $\mathcal{D} := \{(\tau^c_1, \mathcal{T}_{\tau^c_1}), (\tau^c_2, \mathcal{T}_{\tau^c_2}), ..., (\tau^c_n, \mathcal{T}_{\tau^c_n})\}$, where $\tau^c_j$ denote hypothetical scenarios and $\mathcal{T}_{\tau^c_j}$
are the corresponding test cases. For any triple $\tau$, we assign the following two binary labels: $y_{\tau}$ indicates whether $\tau$ is present in the \emph{original fact set} $\mathcal{F}$ and $y^c_{\tau}$ indicates whether $\tau$ belongs to the \emph{fact set of the counterfactual graph} induced by $\tau^c_j$, denoted by $\mathcal{F}_{\tau^c_j}$. 
The prediction for $y^c_{\tau}$ made by a method is denoted by $\widehat{y}^c_{\tau}$.\\
\\
\textbf{F1:}
For this metric, we consider all test cases of all hypothetical scenarios without any restrictions. It gives an indication of the overall predictive performance on counterfactual fact sets. We choose the F1-score due to the imbalanced label distribution of our constructed test cases. The metric is given by
$$\text{F1} = \frac{2tp}{2tp + fn + fp},$$
where 
\begin{align*}
tp &= \sum\limits_{j=1}^{n}\sum\limits_{\tau \in \mathcal{T}_{\tau^c_j}} \mathbb{I}(y^c_{\tau} = 1 \land \widehat{y}^c_{\tau} = 1),\\
fn &= \sum\limits_{j=1}^{n}\sum\limits_{\tau \in \mathcal{T}_{\tau^c_j}} \mathbb{I}(y^c_{\tau} = 1 \land \widehat{y}^c_{\tau} = 0),\\
fp &= \sum\limits_{j=1}^{n}\sum\limits_{\tau \in \mathcal{T}_{\tau^c_j}} \mathbb{I}(y^c_{\tau} = 0 \land \widehat{y}^c_{\tau} = 1)
\end{align*}
\textbf{Changed:}
We denote the set of \emph{changed facts} in $\mathcal{T}_{\tau^c_j}$ by $\mathcal{T}^*_{\tau^c_j}$. Formally, 
\begin{align*}
\mathcal{T}^*_{\tau^c_j} := \{\tau \in \mathcal{T}_{\tau^c_j}: (y_\tau = 0 \,\land\, y^c_\tau = 1) \;\lor\\
(y_\tau = 1 \,\land\, y^c_\tau = 0)\}.
\end{align*}

Intuitively, $\mathcal{T}^*_{\tau^c_j}$ is comprised of facts that were not present in the original graph but emerge in the counterfactual KG or vice versa. We compute the accuracy on these cases with respect to $y_{\tau}^c$.
$$\text{Changed} = \frac{\widetilde{tp} + \widetilde{tn}}{\widetilde{tp} + \widetilde{fn} + \widetilde{fp} + \widetilde{tn}},$$
where 
\begin{align*}
\widetilde{tp} &= \sum\limits_{j=1}^{n}\;\sum\limits_{\tau \in \mathcal{T}^*_{\tau^c_j}} \mathbb{I}(y^c_{\tau} = 1 \land \widehat{y}^c_{\tau} = 1),\\
\widetilde{fn} &= \sum\limits_{j=1}^{n}\;\sum\limits_{\tau \in \mathcal{T}^*_{\tau^c_j}} \mathbb{I}(y^c_{\tau} = 1 \land \widehat{y}^c_{\tau} = 0),\\
\widetilde{fp} &= \sum\limits_{j=1}^{n}\;\sum\limits_{\tau \in \mathcal{T}^*_{\tau^c_j}} \mathbb{I}(y^c_{\tau} = 0 \land \widehat{y}^c_{\tau} = 1),\\
\widetilde{tn} &= \sum\limits_{j=1}^{n}\;\sum\limits_{\tau \in \mathcal{T}^*_{\tau^c_j}} \mathbb{I}(y^c_{\tau} = 0 \land \widehat{y}^c_{\tau} = 0)
\end{align*}

Note that in the case of automatically generated labels (Table \ref{tab:main_table} and \taskname{}-CoDEx-M* (E) in Table \ref{tab:case_study}), $\mathcal{T}^*_{\tau^c_j}$ only consists of \emph{emerging facts} and hence the ground truth labels $y^c_{\tau}$ are always positive.\\
\\
\textbf{Unchanged:}
Let $\overline{\mathcal{T}}_{\tau^c_j}$ denote the set of \emph{unchanged} facts in $\mathcal{T}_{\tau^c_j}$.
Formally,
\begin{align*}
\overline{\mathcal{T}}_{\tau^c_j} := \{\tau \in \mathcal{T}_{\tau^c_j}: (y_\tau = 0 \land y^c_\tau = 0) \;\lor\\ (y_\tau = 1 \land y^c_\tau = 1)\}.
\end{align*}
Intuitively, $\mathcal{T}_{\tau^c_j}$ is comprised of facts that do not change their label between $\mathcal{F}$ and $\mathcal{F}_{\tau^c_j}$. We compute the F1-score on such instances due to their imbalanced label distribution in our constructed test cases.
$$\text{Unchanged} = \frac{2\overline{tp}}{2\overline{tp} + \overline{fn} + \overline{fp}},$$
where 
\begin{align*}
\overline{tp} &= \sum\limits_{j=1}^{n}\sum\limits_{\tau \in \overline{\mathcal{T}}_{\tau^c_j}} \mathbb{I}(y^c_{\tau} = 1 \land \widehat{y}^c_{\tau} = 1),\\
\overline{fn} &= \sum\limits_{j=1}^{n}\sum\limits_{\tau \in \overline{\mathcal{T}}_{\tau^c_j}} \mathbb{I}(y^c_{\tau} = 1 \land \widehat{y}^c_{\tau} = 0),\\
\overline{fp} &= \sum\limits_{j=1}^{n}\sum\limits_{\tau \in \overline{\mathcal{T}}_{\tau^c_j}} \mathbb{I}(y^c_{\tau} = 0 \land \widehat{y}^c_{\tau} = 1)
\end{align*}

\section{CoDEx Resources}\label{sec:codex_resources}
We use the CoDEx knowledge graph completion benchmark, which is comprised of three knowledge graphs (S, M, L) collected from Wikidata based on seed entities and relations for 13 different domains (e.g., media and entertainment, politics, science) \cite{safavi-koutra-2020-codex}.
Table \ref{tab:resources} provides an overview over the resources provided by CoDEx.
\begin{table}
\centering
\resizebox{.47\textwidth}{!}{
\begin{tabular}{lccccccc}\toprule
 & $|\mathcal{E}|$ & $|\mathcal{R}|$ & $|\mathcal{F}_{train}|$ & $|\mathcal{F}_{val}|$ & $|\mathcal{F}_{test}|$ & Negatives\\
\midrule
S & 2034 & 42 & 32888 & 1827 & 1828 & Yes \\
M & 17050 & 51 & 185584 & 10310 & 10311 &  Yes \\
L & 77951 & 69 & 551193 & 30622 & 30622 & No \\\bottomrule
\end{tabular}
}
\caption{Overview of CoDEx datasets \cite{safavi-koutra-2020-codex}. The column ``Negatives'' indicates whether hard, verified validation and test triples are available.}
\label{tab:resources}
\end{table}

\section{Details of Dataset Creation}\label{sec:cckg_appendix}
This section contains details of the \taskname{} dataset creation that were omitted in Section \ref{sec:cckg} due to space constraints and gives a full algorithmic description of the procedure.

\subsection{Formal Description}
Section \ref{sec:cckg} provides a high-level description on how we create \taskname{} test instances based on the first body atom of a rule. This section covers the case where the second body atom is selected for creating the hypothetical scenario and contains formal descriptions of the employed constraints. 

In the following, we define an \emph{atom} variable to distinguish between hypothetical scenarios derived from the first (atom = 1) versus the second atom (atom = 2). 
The general setup is equivalent for both settings:
Given a rule $(r_1, r_2, r_3)$, we search for existing edges $e_1 := (x, r_1, y) \in \mathcal{F}_{train}$ and $e_2 := (\bar{y}, r_2, z) \in \mathcal{F}_{train}$, such that $\tau^i := (x, r_3, z) \notin \mathcal{F}$.
We employ the following constraints I1, I2, and I3 when sampling $e_1$ and $e_2$ to ensure plausible hypothetical scenarios and inferences.\\
\textbf{I1:} if $atom$ = 1: $\exists a \in \mathcal{E}: (a, r_1, \bar{y}) \in \mathcal{F}$,\\
\hspace*{0.5cm} if $atom$ = 2: $\exists b \in \mathcal{E}: (y, r_2, b) \in \mathcal{F}$\\
\textbf{I2:}
$\exists c \in \mathcal{E}: (x, r_3, c) \in \mathcal{F}$\\
\textbf{I3:}
$\exists d \in \mathcal{E}: (d, r_3, z) \in \mathcal{F}$\\
The above constraints ensure that the constructed triples $\tau^c$ and $\tau^i$ have suitable entities for the given relation. Intuitively, I1 ensures that we only select links $(\bar{y}, r_2, z)$ for which the resulting counterfactual triple $(x, r_1, \bar{y})$ is sensible.

When corrupting a given triple $(h, r, t)$, we employ the constraints C1, C2 and C3 when selecting $h'$, $r'$ and $t'$.\\
\textbf{C1:}
$\exists a \in \mathcal{E}: (h', r, a) \in \mathcal{F}$\\
\textbf{C2:}
$\exists b \in \mathcal{E}: (b, r, t') \in \mathcal{F}$\\
\textbf{C3:} $(h', r, t), (h, r, t'), (h, r', t) \notin \mathcal{F} \cup  \mathcal{F}^+_{\Delta}$,\\
where $\mathcal{F}^+_{\Delta}$ denotes the set of inferences made by \emph{all rules} in our rule set, given the hypothetical scenario. C1 and C2 promote challenging head and tail corruptions, which cannot be trivially identified due to the triples being nonsensical. C3 ensures that the generated corruptions are neither present in the original KG nor implied by the given hypothetical scenario, given our rule set. In rare cases, enforcing C1 or C2 would only allow to create triples that are already in the graph or implied by our rules. In such instances, we sample from the full entity set $\mathcal{E}$ instead, while still respecting C3.

\begin{algorithm}
\DontPrintSemicolon
\caption{Creation of \taskname{} instances for a given rule.}\label{alg:cckg}
\KwData{knowledge graph $\mathcal{G} = \{\mathcal{E}, \mathcal{R}, \mathcal{F}\}$,\newline
inference rule $\delta$,\newline
\# of CFs to generate per atom $M$
}
\KwResult{\taskname{} instances for rule $\delta$}
$\mathcal{D}_{\delta} \leftarrow \{\}$\\
\For{$atom \in \{1, 2\}$}{
\For{$n \in \{1, ..., M\}$}{
Randomly sample\\
$(x, r_1, y)$ and $(\bar{y}, r_2, z)$ from $\mathcal{F}_{train}$,\\ according to I1, I2, I3, I4\\
\uIf{$atom = 1$}{
$\tau^c \leftarrow (x, r_1, \bar{y})$
}
\Else{$\tau^c \leftarrow (y, r_2, z)$}
$\tau^i \leftarrow (x, r_3, z)$\\
Sample $\tau^n_1$, $\tau^n_2$ from $\mathcal{N}_1(\tau^c)$\;
Sample $\tau^f$ from $\mathcal{F}\setminus\mathcal{N}_1(\tau^c)$\;
Create corruptions for $\tau^i$, $\tau^n_1$, $\tau^n_2$, $\tau^f$\\according to C1, C2, C3\;
$\mathcal{T}_{\tau^c} \leftarrow \{\tau^i, \tau^n_1, \tau^n_2, \tau^f$, \\\hspace*{1.2cm} $\tau^i_{h'}$, $\tau^i_{r'}$, $\tau^i_{t'}$, ..., $\tau^f_{h'}$, $\tau^f_{r'}$, $\tau^f_{t'}$\}
$\mathcal{D}_{\delta} \leftarrow \mathcal{D}_{\delta} \cup \{(\tau^c, \mathcal{T}_{\tau^c})\}$
}
}
\end{algorithm}

\subsection{Additional Constraints for P361 and P463}\label{sec:additional_constraints}
For counterfactual triples $\tau^c$ using the relation P361 ("part of") or P463 ("member of"), we introduce an additional condition when sampling $e_1$ and $e_2$ based on \emph{entity types} in order to avoid nonsensical hypothetical scenarios, such as (Iraq, part of, The Quarrymen).
Entity types are available for every entity in the CoDEx dataset \cite{safavi-koutra-2020-codex} and provide additional information regarding the entity. For instance, "France" is associated with the entity type "country" (among others) and "7B" is tagged as a "musical group". We denote the set of entity types associated with an entity $e \in \mathcal{E}$ by $type(e)$. We define the following constraint:\\
\textbf{I4:} $type(\bar{y}) \cap type(y) \neq \{\}$, \\\hspace*{0.7cm}if $atom=1$ and $r_1 \in \{P361, P463\}$ or\\\hspace*{0.7cm}if $atom=2$ and $r_2 \in \{P361, P463\}$\\
This condition heuristically ensures that the entity that replaces the original head/tail of a triple to create a hypothetical scenario is of a similar type as the original entity. In the example above, (Iraq, part of, The Quarrymen) is no longer a valid generation when I4 is enforced, since the "The Quarrymen" shares no entity type with the original tail "Middle East".
\subsection{Algorithm}
Algorithm \ref{alg:cckg} describes the dataset creation for \taskname. $\mathcal{N}_1(\tau^c)$ denotes the one-hop neighborhood of $\tau^c$, excluding the context triggering the rule. Note that $\mathcal{N}_1(\tau^c)$ is defined on the full fact set $\mathcal{F}$ = $\mathcal{F}_{train}\, \cup \, \mathcal{F}_{valid} \, \cup \, \mathcal{F}_{test}$. The remaining notation follows Sections \ref{sec:task} and \ref{sec:cckg}.

\section{Human Dataset Verification}\label{sec:human_verification_appendix}
This section details the recruitment of the annotators as well as the guidelines provided to them via the annotation interface.
\subsection{Annotator Recruitment and Demographic}
We recruited annotators on a voluntary basis and did not offer financial compensation. Labelers were made aware that their annotations will be used and published in a scientific paper. We recruited twelve annotators in total, including the authors. All of the annotators have at least a Bachelor's degree. The annotation effort varied between different annotators, with the lowest number of annotated samples being 20 and the highest being 1020.

\subsection{Annotation Guidelines}
This section contains the annotation guidelines provided to the annotators on the annotation interface. Explanations written in \emph{italic} were added during the annotation process as they were requested by annotators. Apart from the guidelines below, the annotators were provided with instructions on how to use the annotation interface.

The main goal of the task is to judge the plausibility of presented statements, given a hypothetical scenario and potentially relevant context.\\
\\
Each annotation prompt presented to you will consist of the following elements:
\begin{itemize}
\item a \textbf{hypothetical scenario}, which you should assume to be true
\item a \textbf{context}, which gives additional information regarding the entities in the scenario
\item a \textbf{statement}, which should be labeled as \textbf{likely}, \textbf{unlikely}, or \textbf{unsure/too little information}
\end{itemize}
Please assign the label \textbf{likely} if you think the presented statement is likely to hold given the hypothetical scenario, the context, and your world knowledge. Assign \textbf{unlikely} if you do not think so. Assign the label \textbf{unsure/too little information} if you cannot confidently judge the plausibility of the statement based on the presented information.\\
\\
Expressions in parentheses denote \textbf{entity types}, which provide additional information for each entity. They can be helpful when reasoning with lesser-known entities. For instance, the entity '7B' is associated with the entity type 'musical group' to clarify that '7B' refers to a band.\\
\\
Each statement follows the general structure 'Is it then plausible that ..., given that this IS NOT the case in the real world?' or 'Is it then STILL plausible that ..., given that this IS the case in the real world?'. Please pay attention to this difference when labeling.\\
\\
\textbf{Example 1:}\\
\\
\textbf{Hypothetical scenario}: Paris (city with millions of inhabitants, city, big city) is located in Japan (island nation, sovereign state, country)\\
\\
\textbf{Context}: Japan (island nation, sovereign state, country) is part of the continent Asia (continent, continental area and surrounding islands)\\
\\
\textbf{Question}: Is it then plausible that Paris (city with millions of inhabitants, city, big city) belongs to the continent Asia (continent, continental area and surrounding islands), given that this is not the case in the real world?\\
\\
\emph{In this scenario, Paris belonging to the continent Asia will likely be the case, hence, we assign the label 'likely'.}
\\\\
\textbf{Example 2:}\\
\\
In some cases, the statement you are presented with might not have a strong, obvious connection to the hypothetical scenario (such as shared entities). This is intended and should not affect your annotation. For instance, you might encounter an example similar to the following:\\
\\
\textbf{Hypothetical scenario}: Paris (city with millions of inhabitants, city, big city) is located in Japan (island nation, sovereign state, country)\\
\\
\textbf{Context}: Japan (island nation, sovereign state, country) is part of the continent Asia (continent, continental area and surrounding islands)\\
\\
\textbf{Question}: Is it then still plausible that English (modern language, natural language, language) is the official language of United Kingdom (country, sovereign state, island nation), given that this is the case in the real world?\\
\\
If you believe that this statement is still plausible in a world where Paris is in Japan, assign 'likely'. If you think otherwise or cannot make a decision based on the presented information, assign 'unlikely' or 'unsure/too little information' respectively. \emph{In the example above, we would expect the label 'likely', since Paris moving to Japan should not affect the official language of the United Kingdom.}\\
\\
\textbf{Example 3:}\\
\\
The statements might not be sensible for all examples. For instance, you could come across a statement like:\\
\\
\textbf{Hypothetical scenario}: Paris (city with millions of inhabitants, city, big city) is located in Japan (island nation, sovereign state, country)\\
\\
\textbf{Context}: Japan (island nation, sovereign state, country) is part of the continent Asia (continent, continental area and surrounding islands)\\
\\
\textbf{Question}: Is it then plausible that Paris (city with millions of inhabitants, city, big city) is the unmarried partner of Asia (continent, continental area and surrounding islands), given that this is not the case in the real world?\\
\\
These examples are intentional and you should annotate them according to the same scheme as the other examples.
\emph{In the example above, we would expect the label 'unlikely', since a city cannot be the unmarried partner of a continent.}

\section{KGE Methods}\label{sec:architectures}
\textbf{TransE} \cite{transe} treats relations as translations in the embedding space. It finds embedding vectors $\mathbf{h}, \mathbf{r}, \mathbf{t} \in \mathbb{R}^{d_e}$ such that $\textbf{h} + \textbf{r} \approx \textbf{t}$ for $(h, r, t) \in \mathcal{F}$, and uses the scoring function $\phi(h, r, t) = -||\mathbf{h} + \mathbf{r} - \mathbf{t}||_2$.
\textbf{ComplEx} \cite{pmlr-v48-trouillon16} maps entities and relations to the complex space and leverages the scoring function $\phi(h, r, t) = \text{Re}(\langle \mathbf{r}, \mathbf{h}, \bar{\mathbf{t}} \rangle)$, where $\mathbf{h}, \mathbf{r}, \mathbf{t} \in \mathbb{C}^{d_e}$ and $\bar{\mathbf{t}}$ denotes the complex conjugate of $\mathbf{t}$. ComplEx is particularly well-suited for modeling antisymmetric relations (e.g., "part of").
\textbf{RESCAL} \cite{rescal} represents the  fact set as a three-dimensional tensor $\mathcal{X}$ with $\mathcal{X}_{i,j,r}=1$ if ($i$, $r$,
$j$) $\in \mathcal{F}$ and $\mathcal{X}_{i,j,r}=0$ otherwise. Representations for entities and relations are obtained via a low-rank factorization $\mathcal{X}_r \approx ER_rE^T$, $E\in \mathbb{R}^{|\mathcal{E}| \times d_e}$, $R_r \in \mathbb{R}^{d_e\times d_e}$ . The score of a given triple is computed as $\phi(h, r, t)=\mathbf{h}^TR_r\mathbf{t}$, where $\mathbf{h}$ and $\mathbf{t}$ are the rows of $E$ corresponding to $h$ and $t$.
Similarly, \textbf{TuckER} \cite{balazevic-etal-2019-tucker} leverages Tucker decomposition \cite{tucker1966some} to find representations $\mathbf{h}, \mathbf{t} \in \mathbb{R}^{d_e}$, $\mathbf{r} \in \mathbb{R}^{d_r}$, as well as a core tensor $\mathcal{W} \in \mathbb{R}^{d_e \times d_r \times d_e}$ which allows sharing knowledge between all entity and relation embeddings. The scores are defined as $\phi(h, r, t) = \mathcal{W} \times_1 \mathbf{h} \times_2 \mathbf{r} \times_3 \mathbf{t}$, where $\times_i$ denotes the tensor product along the $i$-th mode. TuckER was shown to be effective for modeling compositional relations \cite{safavi-koutra-2020-codex}. \textbf{ConvE} \cite{dettmers2018convolutional} is a convolutional architecture described by $\phi(h, r, t) = f(\text{vec}(f([\mathbf{M}_{\mathbf{h}}; \mathbf{M}_{\mathbf{r}}]*\mathbf{\omega}))\mathbf{W})\mathbf{t}$, where $\mathbf{M}_{\mathbf{h}}$ and $\mathbf{M}_{\mathbf{r}}$ are 2D-reshapings of entity and relation embeddings, $\mathbf{\omega}$ describes the convolutional filters, and 
$\text{vec}$ denotes vectorization \cite{ji2021survey}.

\section{Experimental Setting}
\subsection{Implementation and Runtime Details} \label{sec:hardwear}
We run our experiments on a single Tesla V100 GPU with 16GB of memory on a Nvidia DGX1 server. COULDD hyperparameter tuning takes between around 35 minutes and 65 minutes and a single run on the test set takes between 3 and 20 minutes, depending on the model and dataset.

For KGE embeddings, we use the pre-trained CoDEx models \cite{safavi-koutra-2020-codex}, which were trained using LibKGE \cite{libkge}. For our experiments with COULDD, we slightly adapt the LibKGE implementation to allow for our proposed training scheme. For hyperparameter tuning, we use the \texttt{GridSampler} implemented in Optuna \cite{optuna} (version 3.3.0). For computing performance metrics (F1, accuracy, confusion matrix), we use scikit-learn (version 1.3.0).
All results are reproducible with seed 0.

\subsection{Hyperparameters}\label{sec:hyperparameters_app}
 Table~\ref{tab:hyperparameter} lists the hyperparameters used for our experiments \ref{tab:main_table}. 
 Bold parameters were tuned for COULDD on a validation set via grid search, while the remaining parameters were carried over from the pre-trained models provided by \citet{safavi-koutra-2020-codex}. For further details on the pre-trained models, please refer to \citet{safavi-koutra-2020-codex}. The learning rate ($\alpha$) was tuned in the range of \{0.001, 0.01, 0.1, 0.15, 0.2\}. The number of additional samples ($N$) was chosen in the range of \{0, 127, 255, 511, 1023\} for all models except ConvE. For ConvE, the range was reduced to \\\{127, 255, 511, 1023\} because of its BatchNorm layer.

\subsection{ChatGPT Experimental Setup}\label{sec:appendix_gpt}

For our experiments with ChatGPT, we used the OpenAI API. We used the model \textit{gpt-3.5-turbo-0613} and set the temperature to 0 for all experiments.  
The given system prompt, prompt templates, as well as an input and output example are given in Table \ref{tab:appendix_chatgpt_prompts}.

For two inputs in \taskname-CoDEx-S, 12 in \taskname-CoDEx-M and 23 in \taskname-CoDEx-L, ChatGPT did not answer in the desired format. We nevertheless attempted to extract the answer using a regular expression but this process could potentially be error-prone. For one instance of \taskname-CoDEx-M, ChatGPT gave the answer ``inconclusive'', which is not one of our accepted labels. We counted this instance as wrongly classified in our experiments.

\section{Evaluation per Test Type}\label{sec:eval_per_test_type}
Table \ref{tab:appendix_type_table} provides the performance per test case for the results in Table \ref{tab:main_table}. The results suggest that head corruptions of valid facts are generally harder to identify than tail corruptions. This is likely partially due to the setup of the CoDEx triple classification benchmark, which tunes decision thresholds solely on tail corruptions and always uses object-oriented scoring, even when reciprocal relations are available. We adopted this setup to make our results comparable to the original CoDEx paper \cite{safavi-koutra-2020-codex}.

\section{Case Study on CoDEx-M}
\subsection{\taskname{}-CoDEx-M* Label Distribution}\label{sec:cckg_m_star_stats}
Table \ref{tab:cckg_m_star_dist} gives the label distribution of expected labels (according to our assumptions) and majority vote labels on \taskname{}-CoDEx-M*.
\begin{table}[!ht]
\centering
\begin{tabular}{lcccc}\toprule
& \multicolumn{2}{c}{Expected (E)} & \multicolumn{2}{c}{Human (H)}\\\toprule
 & 0 & 1 & 0 & 1\\\midrule
$\tau^i$ & 0 & 181 & 33 & 148\\
$\tau^f$ & 0 & 255 & 0 & 255\\
$\tau^n$ & 0 & 495 & 14 & 481\\
$\tau^i_{r'}$ & 228 & 0 & 214 & 14\\\bottomrule
\end{tabular}
\caption{Label distribution in the \taskname{}-CoDEx-M* test set with expected labels (E) and human-assigned (H) labels.}
\label{tab:cckg_m_star_dist}
\end{table}

\subsection{Rule-wise Performance on Filtered CoDEx-M}\label{sec:inferential_appendix}
In Section \ref{sec:inferential_case_study}, we investigate how well the pre-trained CoDEx models can infer CoDEx-M test triples that are implied by AMIE3 rules. Note that these experiments do not introduce any hypothetical scenarios. Table \ref{tab:inferential_rule_wise} provides information about the performance on individual rules. Note that a triple can potentially be inferred by multiple rules and hence contribute to the rule-wise performance with respect to multiple rules.

\subsection{Confusion matrix on \taskname{}-CoDEx-M*}
Table \ref{tab:perf_analysis} gives the confusion matrix for COULDD and ChatGPT on \taskname{}-CoDEx-M* with manually assigned labels.
 \begin{table*}[!ht]
  \centering
 \resizebox{\textwidth}{!}{
 \begin{tabular}{lrrrrr}\toprule
 & RESCAL & TransE & ComplEx & ConvE & TuckER\\\hline
\textbf{\taskname{}-CoDEx-S} & & & & &\\\hline
 Embedding size & 512 & 512 & 512 & 256 & 512\\
  Reciprocal & No & Yes & Yes & Yes & Yes\\ 
 Optimizer & Adagrad & Adagrad & Adam & Adagrad & Adagrad\\
 Regularization & & & & &\\
 \hspace{0.5cm} \small{Type} & $l_3$ & $l_2$ & None & $l_3$ & $l_1$\\
\hspace{0.5cm} \small{Entity embeddings} & $2.18 \times 10^{-10}$ & $1.32 \times 10^{-7}$ & $9.58 \times 10^{-13}$ & $3.11 \times 10^{-15}$ & $3.47 \times 10^{-15}$\\
\hspace{0.5cm} \small{Relation embeddings} & $3.37 \times 10^{-14}$ & $3.72 \times 10^{-18}$ & 0.0229 & $4.68 \times 10^{-9}$ & $3.43 \times 10^{-14}$\\
\hspace{0.5cm} \small{Frequency weighting} & False & False & True & True & True\\
Dropout & & & & &\\
\hspace{0.5cm} \small{Entity embeddings} & 0.0 & 0.0 & 0.0793 & 0.0 & 0.1895\\
\hspace{0.5cm} \small{Relation embeddings} & 0.0804 & 0.0 & 0.0564 & 0.0 & 0.0\\
\hspace{0.5cm} \small{Feature map (ConvE)} & - & - & - & 0.2062 & -\\
\hspace{0.5cm} \small{Projection (ConvE)} & - & - & - & 0.1709 & -\\
\textbf{Additional samples ($N$)} & 127 & 255 & 127 & 255 & 255\\
\textbf{Learning rate ($\alpha$)} & 0.01 & 0.01 & 0.1 & 0.001 & 0.01\\\hline
\textbf{\taskname{}-CoDEx-M} & & & & &\\\hline
 Embedding size & 256 & 512 & 512 & 512 & 512\\
 Reciprocal & Yes & Yes & Yes & Yes & Yes\\ 
 Optimizer & Adagrad & Adagrad & Adagrad & Adagrad & Adagrad\\
 Regularization & & & & &\\
 \hspace{0.5cm} \small{Type} & $l_2$ & $l_2$ & $l_3$ & $l_1$ & $l_1$\\
\hspace{0.5cm} \small{Entity embeddings} &
$9.56 \times 10^{-7}$ &
$1.32 \times 10^{-7}$ &
$1.34 \times 10^{-10}$ &
$1.37 \times 10^{-10}$ &
$3.47 \times 10^{-15}$\\
\hspace{0.5cm} \small{Relation embeddings} &
$2.56 \times 10^{-17}$ &
$3.72 \times 10^{-18}$ &
$6.38 \times 10^{-16}$ &
$4.72 \times 10^{-10}$ &
$3.4 \times 10^{-14}$\\
\hspace{0.5cm} \small{Frequency weighting} & False & False & True & True & True\\
Dropout & & & & &\\
\hspace{0.5cm} \small{Entity embeddings} & 0.0 & 0.0 & 0.1196 & 0.0 & 0.1895\\
\hspace{0.5cm} \small{Relation embeddings} & 0.0 & 0.0 & 0.3602 & 0.0348 & 0.0\\
\hspace{0.5cm} \small{Feature map (ConvE)} & - & - & - & 0.3042 & -\\
\hspace{0.5cm} \small{Projection (ConvE)} & - & - & - & 0.2343 & -\\
\textbf{Additional samples ($N$)} & 255 & 511 & 0 & 511 & 1023\\
\textbf{Learning rate ($\alpha$)} & 0.01 & 0.01 & 0.1 & 0.001 & 0.01\\\hline
\textbf{\taskname{}-CoDEx-L} & & & & &\\\hline
 Embedding size & 128 & 128 & 128 & 256 & 256\\
Reciprocal & No & Yes & Yes & Yes & No\\ 
 Optimizer & Adagrad & Adam & Adagrad & Adagrad & Adagrad\\
 Regularization & & & & &\\
 \hspace{0.5cm}\small{Type} & $l_2$ & $l_2$ & $l_2$ & $l_1$ & $l_2$\\
\hspace{0.5cm} \small{Entity embeddings} &
$2.01 \times 10^{-16}$ &
$7.98 \times 10^{-14}$ &
$2.01 \times 10^{-16}$ &
$6.10 \times 10^{-16}$ &
$8.06 \times 10^{-11}$\\
\hspace{0.5cm} \small{Relation embeddings} &
$3.52 \times 10^{-13}$ &
$3.42 \times 10^{-9}$ &
$3.52 \times 10^{-13}$ &
$1.03 \times 10^{-16}$ &
$7.19 \times 10^{-19}$\\
\hspace{0.5cm} \small{Frequency weighting} & True & False & True & True & True\\
Dropout & & & & &\\
\hspace{0.5cm} \small{Entity embeddings} & 0.0 & 0.0 & 0.0 & 0.0064 & 0.1606\\
\hspace{0.5cm} \small{Relation embeddings} & 0.0 & 0.0 & 0.0 & 0.0 & 0.0857\\
\hspace{0.5cm} \small{Feature map (ConvE)} & - & - & - & 0.1530 & -\\
\hspace{0.5cm} \small{Projection (ConvE)} & - & - & - & 0.4192 & -\\
\textbf{Additional samples ($N$)} & 0 & 1023 & 0 & 127 & 127\\
\textbf{Learning rate ($\alpha$)} & 0.1 & 0.2 & 0.2 & 0.01 & 0.01\\\bottomrule
\end{tabular}
}
\caption{Hyperparameters for COULDD experiments. Bold hyperparameters were tuned by us. The remaining are taken from the original CoDEx paper \cite{safavi-koutra-2020-codex} and kept the same for our experiments.}\label{tab:hyperparameter}
\end{table*}
\begin{table*}
    \centering
    \resizebox{0.9\textwidth}{!}{
    \begin{tabular}{l|p{13cm}}
    \toprule
     System Prompt & You are a helpful, honest data labeler who classifies the plausibility of a hypothetical scenario. \\
 & You will be presented: \\
& Hypothetical scenario: This fact is wrong in the real world, but you assume it is true in the current hypothetical world. \\
& Context:  It is true in the hypothetical world, and gives additional information you can use to reason about the plausibility of the presented statement \\
& A statement which is supposed to be labeled as 'likely' or 'unlikely'.
\\
& \\
& Each statement follows the general structure 'Is it then plausible that ..., given that this IS NOT the case in the real world?' or 'Is it then STILL plausible that ..., given that this IS the case in the real world?'. Please pay attention to this difference when labeling. \\
\midrule
     Statement Template    & \{\{ Statement \}\} \\
     & \\
     & Return 'likely' if you think the presented statement is likely to hold given the hypothetical scenario, the context, and your world knowledge. \\
 & - If a fact was "the case" in the real world, respond 'likely' if the hypothetical scenario does not make it implausible \\
 & - If a fact was "not the case" in the real world, respond 'likely' if the hypothetical scenario makes it plausible. \\
& Return 'unlikely' only if the presented statement is unlikely to hold given the hypothetical scenario, the context, and your world knowledge. \\
 & - If a fact was "the case" in the real world, respond 'unlikely' if the hypothetical scenario makes it implausible. \\
 & - If a fact was "not the case" in the real world, respond 'unlikely' if the hypothetical scenario does not make it plausible. \\
 & \\
& If The hypothetical scenario and the context do not provide any specific information for the statement, respond 'unlikely' given that the statement is "not the case in the real world" and 'likely' if the statement "is the case in the real world". \\
& \\
& Respond with a JSON. First, include a explanation of your reasoning in the 'explanation' field and give then an answer in the 'answer' field. \\
\midrule
Example Statement & Hypothetical scenario: Julio Iglesias (human) is married to Rachel Weisz (human) \\
& \\
& Context: Rachel Weisz (human) is a citizen of United Kingdom (island nation, Commonwealth realm, sovereign state) \\
 & \\
& Is it then plausible that Julio Iglesias (human) is a citizen of United Kingdom (colonial power, island nation, country), given that this is not the case in the real world? \\
\midrule
Original Label & Unlikely \\
\midrule
Expected Label (in CF KG) & Likely \\
\midrule
Example Output & \{ \\
  & "explanation": "Based on the hypothetical scenario and the context provided, it is plausible that Julio Iglesias (human) is a citizen of United Kingdom (colonial power, island nation, country). The context states that Rachel Weisz (human) is a citizen of United Kingdom, which implies that she has citizenship in the country. Since Julio Iglesias is married to Rachel Weisz in the hypothetical scenario, it is likely that he would also have acquired citizenship in the United Kingdom through marriage.", \\
  & "answer": "likely" \\
 & \} \\
    \bottomrule
    \end{tabular}
    }
    \caption{The table shows how ChatGPT was used. It includes templates and an example scenario with the provided output. }
    \label{tab:appendix_chatgpt_prompts}
\end{table*}
\begin{table*}[t]
\begin{center}
\resizebox{\textwidth}{!}{
\begin{tabular}{llllllllllllll}
\toprule
 &  & $\tau^i$ & $\tau^f$ & $\tau^i_{h'}$ & $\tau^f_{h'}$ & $\tau^n_{h'}$ & $\tau^n$ & $\tau^i_{r'}$ & $\tau^f_{r'}$ & $\tau^n_{r'}$ & $\tau^i_{t'}$ & $\tau^f_{t'}$ & $\tau^n_{t'}$ \\
Dataset & Method &  &  &  &  &  &  &  &  &  &  &  &  \\
\midrule
\multirow[t]{10}{*}{CFKGR-CoDEx-S} & RESCAL & 27.12 & \textbf{99.46} & 73.78 & 39.24 & \textbf{56.6} & \textbf{98.55} & \textbf{94.94} & 94.21 & \textbf{94.76} & 79.57 & 51.54 & 61.66 \\
 & COULDD-RESCAL & \textbf{32.48 $\pm$ 0.73} & 99.28 $\pm$ 0.2 & \textbf{73.82 $\pm$ 1.04} & \textbf{43.15 $\pm$ 1.19} & 55.5 $\pm$ 0.53 & 98.17 $\pm$ 0.16 & 94.86 $\pm$ 0.25 & \textbf{94.5 $\pm$ 0.25} & 94.47 $\pm$ 0.11 & \textbf{79.78 $\pm$ 0.56} & \textbf{54.32 $\pm$ 0.41} & \textbf{61.95 $\pm$ 0.16} \\
 & TransE & 23.15 & \textbf{95.84} & 78.84 & 32.91 & 55.70 & \textbf{90.78} & \textbf{97.11} & 95.48 & 93.94 & 86.80 & 52.80 & 68.72 \\
 & COULDD-TransE & \textbf{26.8 $\pm$ 0.81} & 94.39 $\pm$ 0.26 & \textbf{82.78 $\pm$ 0.29} & \textbf{36.56 $\pm$ 0.31} & \textbf{58.16 $\pm$ 0.21} & 89.84 $\pm$ 0.21 & 97.11 $\pm$ 0.16 & \textbf{95.66 $\pm$ 0.11} & \textbf{93.96 $\pm$ 0.04} & \textbf{88.28 $\pm$ 0.81} & \textbf{58.37 $\pm$ 0.27} & \textbf{71.57 $\pm$ 0.07} \\
 & ComplEx & 29.11 & \textbf{98.73} & 72.15 & 44.12 & 58.14 & \textbf{98.82} & \textbf{96.2} & \textbf{97.11} & 96.20 & \textbf{79.39} & 56.06 & 64.65 \\
 & COULDD-ComplEx & \textbf{37.94 $\pm$ 0.67} & 93.31 $\pm$ 0.66 & \textbf{84.27 $\pm$ 0.98} & \textbf{63.83 $\pm$ 0.68} & \textbf{71.32 $\pm$ 0.85} & 94.27 $\pm$ 0.19 & 95.77 $\pm$ 0.14 & 97.0 $\pm$ 0.14 & \textbf{96.42 $\pm$ 0.11} & 79.06 $\pm$ 1.52 & \textbf{72.12 $\pm$ 1.3} & \textbf{74.21 $\pm$ 0.66} \\
 & ConvE & 16.64 & \textbf{97.65} & 81.19 & 43.76 & \textbf{65.46} & \textbf{93.67} & 96.56 & 91.32 & \textbf{87.79} & 92.95 & 53.53 & 73.78 \\
 & COULDD-ConvE & \textbf{16.96 $\pm$ 0.72} & 97.22 $\pm$ 0.18 & \textbf{82.21 $\pm$ 0.18} & \textbf{45.53 $\pm$ 0.4} & 65.23 $\pm$ 0.41 & 93.49 $\pm$ 0.27 & \textbf{96.6 $\pm$ 0.07} & \textbf{91.72 $\pm$ 0.07} & 87.58 $\pm$ 0.17 & \textbf{93.56 $\pm$ 0.18} & \textbf{55.15 $\pm$ 0.49} & \textbf{75.8 $\pm$ 0.15} \\
 & TuckER & 15.01 & \textbf{98.37} & \textbf{83.73} & 45.39 & \textbf{71.34} & 98.55 & \textbf{95.3} & \textbf{96.93} & \textbf{94.21} & \textbf{89.33} & 54.79 & 73.69 \\
 & COULDD-TuckER & \textbf{35.99 $\pm$ 1.0} & 97.72 $\pm$ 0.54 & 78.23 $\pm$ 0.29 & \textbf{48.1 $\pm$ 1.23} & 67.09 $\pm$ 0.56 & \textbf{98.61 $\pm$ 0.07} & 94.29 $\pm$ 0.22 & 96.93 $\pm$ 0.23 & 93.35 $\pm$ 0.15 & 83.11 $\pm$ 0.63 & \textbf{58.59 $\pm$ 0.62} & \textbf{74.23 $\pm$ 0.3} \\
\cline{1-14}
\multirow[t]{10}{*}{CFKGR-CoDEx-M} & RESCAL & 21.57 & \textbf{97.96} & 79.41 & 46.90 & 68.10 & \textbf{95.75} & 91.18 & 91.42 & 91.30 & 87.01 & 58.50 & 75.57 \\
 & COULDD-RESCAL & \textbf{26.23 $\pm$ 0.16} & 96.44 $\pm$ 0.19 & \textbf{81.44 $\pm$ 0.14} & \textbf{48.91 $\pm$ 0.25} & \textbf{70.14 $\pm$ 0.27} & 94.49 $\pm$ 0.19 & \textbf{91.23 $\pm$ 0.08} & \textbf{91.47 $\pm$ 0.23} & \textbf{91.31 $\pm$ 0.09} & \textbf{87.19 $\pm$ 0.16} & \textbf{59.54 $\pm$ 0.32} & \textbf{76.41 $\pm$ 0.12} \\
 & TransE & 23.61 & \textbf{88.56} & 76.31 & 36.11 & 62.50 & \textbf{75.37} & \textbf{92.97} & 92.89 & 89.26 & 86.19 & 53.84 & 70.14 \\
 & COULDD-TransE & \textbf{26.06 $\pm$ 0.25} & 85.85 $\pm$ 0.18 & \textbf{76.83 $\pm$ 0.27} & \textbf{38.94 $\pm$ 0.18} & \textbf{63.68 $\pm$ 0.16} & 74.31 $\pm$ 0.14 & 92.78 $\pm$ 0.1 & \textbf{93.17 $\pm$ 0.04} & \textbf{89.31 $\pm$ 0.03} & \textbf{86.75 $\pm$ 0.14} & \textbf{57.63 $\pm$ 0.27} & \textbf{70.92 $\pm$ 0.09} \\
 & ComplEx & 11.60 & \textbf{97.96} & \textbf{89.38} & \textbf{49.02} & \textbf{75.08} & \textbf{97.55} & \textbf{93.63} & \textbf{94.61} & \textbf{92.65} & \textbf{94.69} & \textbf{59.56} & \textbf{80.39} \\
 & COULDD-ComplEx & \textbf{34.67 $\pm$ 0.23} & 97.96 $\pm$ 0.0 & 79.17 $\pm$ 0.53 & 48.95 $\pm$ 0.08 & 69.58 $\pm$ 0.21 & 97.21 $\pm$ 0.1 & 93.09 $\pm$ 0.11 & 94.59 $\pm$ 0.03 & 92.34 $\pm$ 0.07 & 90.36 $\pm$ 0.31 & 59.48 $\pm$ 0.05 & 79.01 $\pm$ 0.09 \\
 & ConvE & 13.15 & \textbf{93.06} & \textbf{87.09} & 41.91 & 67.97 & \textbf{81.78} & \textbf{95.1} & \textbf{88.32} & \textbf{84.76} & \textbf{94.12} & 53.68 & 77.33 \\
 & COULDD-ConvE & \textbf{17.04 $\pm$ 0.16} & 84.72 $\pm$ 0.45 & 85.38 $\pm$ 0.25 & \textbf{43.94 $\pm$ 0.1} & \textbf{68.84 $\pm$ 0.42} & 71.49 $\pm$ 0.35 & 92.4 $\pm$ 0.14 & 86.18 $\pm$ 0.61 & 81.66 $\pm$ 0.4 & 93.5 $\pm$ 0.4 & \textbf{54.31 $\pm$ 0.59} & \textbf{79.52 $\pm$ 0.17} \\
 & TuckER & 13.15 & 97.96 & \textbf{88.4} & \textbf{50.74} & \textbf{76.76} & 97.14 & \textbf{92.48} & \textbf{91.18} & \textbf{88.77} & \textbf{95.02} & \textbf{58.33} & \textbf{80.8} \\
 & COULDD-TuckER & \textbf{43.69 $\pm$ 0.38} & \textbf{98.33 $\pm$ 0.11} & 73.14 $\pm$ 0.54 & 44.07 $\pm$ 0.58 & 67.06 $\pm$ 0.18 & \textbf{97.99 $\pm$ 0.11} & 91.99 $\pm$ 0.32 & 90.87 $\pm$ 0.46 & 87.57 $\pm$ 0.11 & 90.1 $\pm$ 0.38 & 58.27 $\pm$ 0.68 & 78.7 $\pm$ 0.15 \\
\cline{1-14}
\multirow[t]{10}{*}{CFKGR-CoDEx-L} & RESCAL & 71.47 & \textbf{99.89} & 32.09 & 18.09 & \textbf{23.39} & \textbf{99.63} & 68.92 & 74.08 & \textbf{72.88} & \textbf{51.52} & \textbf{53.91} & 51.41 \\
 & COULDD-RESCAL & \textbf{84.56 $\pm$ 0.35} & 99.89 $\pm$ 0.0 & \textbf{32.37 $\pm$ 0.58} & \textbf{18.16 $\pm$ 0.07} & 23.15 $\pm$ 0.14 & 95.58 $\pm$ 0.21 & \textbf{69.2 $\pm$ 0.48} & \textbf{74.09 $\pm$ 0.07} & 69.94 $\pm$ 0.23 & 45.71 $\pm$ 0.54 & 53.87 $\pm$ 0.07 & \textbf{51.61 $\pm$ 0.21} \\
 & TransE & 66.31 & \textbf{99.41} & 30.07 & 18.31 & 20.68 & 99.25 & \textbf{79.4} & 48.00 & 40.82 & 48.96 & 46.89 & 44.97 \\
 & COULDD-TransE & \textbf{76.56 $\pm$ 0.25} & 98.99 $\pm$ 0.1 & \textbf{30.47 $\pm$ 0.55} & \textbf{27.87 $\pm$ 0.66} & \textbf{22.47 $\pm$ 0.21} & \textbf{99.35 $\pm$ 0.06} & 77.54 $\pm$ 0.18 & \textbf{53.22 $\pm$ 0.35} & \textbf{43.67 $\pm$ 0.19} & \textbf{58.12 $\pm$ 0.3} & \textbf{60.04 $\pm$ 0.82} & \textbf{55.93 $\pm$ 0.37} \\
 & ComplEx & 65.51 & \textbf{99.57} & \textbf{36.14} & 27.25 & \textbf{33.02} & 99.44 & \textbf{90.47} & \textbf{84.62} & \textbf{83.93} & \textbf{58.91} & \textbf{64.93} & \textbf{61.07} \\
 & COULDD-ComplEx & \textbf{82.95 $\pm$ 0.26} & 99.57 $\pm$ 0.0 & 31.73 $\pm$ 0.13 & \textbf{27.29 $\pm$ 0.04} & 29.44 $\pm$ 0.09 & \textbf{99.53 $\pm$ 0.04} & 89.03 $\pm$ 0.12 & 84.57 $\pm$ 0.03 & 83.25 $\pm$ 0.09 & 55.5 $\pm$ 0.29 & 64.84 $\pm$ 0.07 & 59.68 $\pm$ 0.14 \\
 & ConvE & \textbf{61.84} & \textbf{99.52} & 41.35 & 36.46 & 40.66 & \textbf{99.18} & 91.06 & 61.63 & 53.54 & 61.58 & 63.70 & 60.32 \\
 & COULDD-ConvE & 45.53 $\pm$ 0.61 & 94.5 $\pm$ 0.36 & \textbf{61.25 $\pm$ 0.59} & \textbf{53.79 $\pm$ 0.3} & \textbf{57.2 $\pm$ 0.47} & 95.32 $\pm$ 0.18 & \textbf{93.18 $\pm$ 0.31} & \textbf{73.72 $\pm$ 0.73} & \textbf{67.2 $\pm$ 0.54} & \textbf{79.45 $\pm$ 0.36} & \textbf{78.11 $\pm$ 0.42} & \textbf{75.53 $\pm$ 0.51} \\
 & TuckER & 76.74 & \textbf{99.79} & \textbf{27.46} & 14.74 & \textbf{22.78} & 99.65 & \textbf{75.36} & 63.23 & \textbf{62.03} & \textbf{53.33} & 50.56 & \textbf{49.92} \\
 & COULDD-TuckER & \textbf{88.47 $\pm$ 0.34} & 99.74 $\pm$ 0.07 & 25.92 $\pm$ 0.52 & \textbf{15.94 $\pm$ 0.23} & 19.58 $\pm$ 0.29 & \textbf{99.68 $\pm$ 0.04} & 72.59 $\pm$ 0.33 & \textbf{64.14 $\pm$ 0.44} & 61.29 $\pm$ 0.21 & 50.07 $\pm$ 0.29 & \textbf{52.84 $\pm$ 0.14} & 48.13 $\pm$ 0.54 \\
\bottomrule
\end{tabular}
}
\end{center}
\caption{Accuracy by test type of pre-trained embeddings and COULDD on \taskname. For COULDD, we report the mean and standard deviation across 5 runs.}\label{tab:appendix_type_table}
\end{table*}

\begin{table*}[ht!]
    \centering
    \resizebox{\textwidth}{!}{
\begin{tabular}{llll|lllll}
\toprule
& Support & PCA & \# Test & RESCAL & TransE & ComplEx & ConvE & TuckER \\
\midrule
(P112, P27, P17) & 38 & 0.826 & 5 & 1.000 & 1.000 & 0.400 & 1.000 & 0.800 \\
(P20, P37, P1412) & 836 & 0.818 & 36 & 0.972 & 0.972 & 1.000 & 0.944 & 1.000 \\
(P19, P37, P1412) & 665 & 0.790 & 23 & 1.000 & 0.826 & 1.000 & 0.826 & 0.957 \\
(P26, P27, P27) & 682 & 0.661 & 15 & 0.933 & 0.933 & 0.867 & 0.733 & 0.933 \\
(P27, P37, P1412) & 9937 & 0.543 & 416 & 0.962 & 0.918 & 0.993 & 0.901 & 0.978 \\
(P17, P30, P30) & 100 & 0.427 & 5 & 0.200 & 0.000 & 0.200 & 0.000 & 0.600 \\
(P161, P27, P495) & 1464 & 0.406 & 87 & 0.805 & 0.943 & 0.920 & 0.931 & 0.931 \\
(P159, P17, P17) & 137 & 0.346 & 6 & 1.000 & 0.833 & 1.000 & 0.833 & 0.833 \\
(P131, P17, P17) & 82 & 0.297 & 5 & 0.600 & 0.600 & 0.400 & 0.800 & 1.000 \\
(P161, P20, P840) & 87 & 0.134 & 6 & 1.000 & 1.000 & 1.000 & 1.000 & 1.000 \\
\bottomrule
\end{tabular}
}
\caption{Rule-wise performance on the filtered test set of CoDEx-M (see Table \ref{tab:case_study}). For each rule, we report the number of positive examples (``Support'') and PCA confidence (``PCA'') as computed by Amie3 on the full KG and the number of inferences in the filtered test set (``\# Test''). The PCA confidence and number of positive examples are taken from \url{https://github.com/tsafavi/codex}.}\label{tab:inferential_rule_wise}
\end{table*}
\begin{table*}
\resizebox{\textwidth}{!}{
\begin{tabular}{l|cccc|cccc|cccc|cccc}\toprule
 & \multicolumn{16}{c}{\taskname-CoDEx-M* (H)}\\\midrule
& \multicolumn{4}{c|}{$\tau^i$} & \multicolumn{4}{c|}{$\tau^f$} & \multicolumn{4}{c|}{$\tau^n$} & \multicolumn{4}{c}{$\tau_{r'}$}\\
& TN & FP & FN & TP & TN & FP & FN & TP & TN & FP & FN & TP & TN & FP & FN & TP\\\hline
COULDD-RESCAL & 13.2 & 19.8 & 122.4 & 25.6 & 0.0 & 0.0 & 8.2 & 246.8 & 0.0 & 14.0 & 24.0 & 457.0 & 196.8 & 17.2 & 11.0 & 3.0 \\
COULDD-TransE & 17.0 & 16.0 & 121.6 & 26.4 & 0.0 & 0.0 & 31.6 & 223.4 & 5.4 & 8.6 & 116.8 & 364.2 & 201.8 & 12.2 & 12.0 & 2.0 \\
COULDD-ComplEx & 15.4 & 17.6 & 98.0 & 50.0 & 0.0 & 0.0 & 4.0 & 251.0 & 2.0 & 12.0 & 7.4 & 473.6 & 202.4 & 11.6 & 13.4 & 0.6 \\
COULDD-ConvE & 18.2 & 14.8 & 132.6 & 15.4 & 0.0 & 0.0 & 35.2 & 219.8 & 4.2 & 9.8 & 128.4 & 352.6 & 197.2 & 16.8 & 11.8 & 2.2 \\
COULDD-TuckER & 13.2 & 19.8 & 89.2 & 58.8 & 0.0 & 0.0 & 3.4 & 251.6 & 0.0 & 14.0 & 8.2 & 472.8 & 203.8 & 10.2 & 12.0 & 2.0 \\\hline
gpt-3.5-turbo & 21 & 12 & 64 & 84 & 0 & 0 & 188 & 67 & 8 & 6 & 169 & 312 & 125 & 89 & 11 & 3\\
\midrule
\end{tabular}
}
\caption{Performance analysis per test type on \taskname-CoDEx-M* with human-assigned labels. For COULDD, the reported values are averaged over 5 model runs.}\label{tab:perf_analysis}
\end{table*}
\section{Further ChatGPT Observations}\label{sec:chatgpt_observations}
By analyzing the explanations provided by ChatGPT, we found some frequent errors in its responses, which might be indicative of its poor performance on unchanged facts. 

First of all, despite the prompt for near and far facts explicitly stating that the statement ``is the case in the real world'', ChatGPT oftentimes directly references that it was ``not the case in the real world'' in its explanations. This observation is however not consistent across examples and in many instances, ChatGPT explained correctly that it should decide whether a triple is ``still plausible'' given that it ``was the case in the real world''.
This inconsistency could potentially stem from our system prompt (see Appendix \ref{sec:appendix_gpt}), which explains the distinction between the two cases.

Moreover, we noticed that ChatGPT sometimes gave a wrong prediction when the same entity was associated with different entity types in the scenario, context and statement due to random sampling (e.g., Budapest (town in Hungary, capital, enclave) compared to Budapest (city with millions of inhabitants, town in Hungary, enclave)). Keeping entity types consistent could likely help to boost the performance of ChatGPT on our benchmark. Judging from the annotation results in Table \ref{tab:annotation_results}, humans were still largely able to reliably judge the validity of the given statements, despite the varying entity types.

\end{document}